\documentclass[10pt,twocolumn,letterpaper]{article}

\usepackage{iccv}
\makeatletter
\@namedef{ver@everyshi.sty}{}
\makeatother

\usepackage{times}
\usepackage{epsfig}
\usepackage{graphicx}
\usepackage{amsmath}
\usepackage{amssymb}

\usepackage[pagebackref=true,breaklinks=true,colorlinks,bookmarks=false]{hyperref}

\iccvfinalcopy %

\usepackage[breaklinks=true,bookmarks=false]{hyperref}
\iccvfinalcopy %

\def\iccvPaperID{1054} %

\ificcvfinal\fi

\usepackage{times}
\usepackage{booktabs}

\usepackage[capitalize]{cleveref}
\crefname{section}{Sec.}{Secs.}
\Crefname{section}{Section}{Sections}
\Crefname{table}{Table}{Tables}
\crefname{table}{Tab.}{Tabs.}

\usepackage{import}
\usepackage{epsfig}
\usepackage{graphicx}
\usepackage{multirow}
\usepackage{amsmath}
\usepackage{amssymb}
\usepackage{pifont}
\usepackage{textcomp}
\usepackage{tikz}
\usepackage{comment}
\usepackage{color}
\usepackage{xspace}
\usepackage{placeins}
\usepackage{bookmark}
\usepackage{rotating}
\usepackage{tablefootnote}

\usepackage{dsfont}
\usepackage{bm}

\usepackage{colortbl}
\usepackage{enumitem}
\usepackage{footnote}
\usepackage{url}
\usepackage{wrapfig}

\usepackage{makecell}
\usepackage{lipsum}
\usepackage{pgffor}

\definecolor{cyan}{cmyk}{.3,0,0,0}
\definecolor{lightblue}{HTML}{B0C4DE}
\definecolor{darkblue}{HTML}{177CB0}
\newcounter{rownumbers}

\usepackage{float}
\usepackage[ruled]{algorithm2e}

\SetCommentSty{mycommfont}

\usepackage{bbm}
\usepackage{soul}

\usepackage{tikz,siunitx}
\usetikzlibrary{fit}
\usetikzlibrary{shapes.geometric,shapes.symbols}

\renewcommand{\Sigma}{\mathfrak{S}}

\def\1{\bm{1}}

\DeclareMathAlphabet{\mathsfit}{\encodingdefault}{\sfdefault}{m}{sl}
\SetMathAlphabet{\mathsfit}{bold}{\encodingdefault}{\sfdefault}{bx}{n}

\newlength\secmargin
\newlength\paramargin
\newlength\abovetabcapmargin
\newlength\belowtabcapmargin
\newlength\abovefigcapmargin
\newlength\belowfigcapmargin

\makeatletter\renewcommand\paragraph{\@startsection{paragraph}{4}{\z@}
  {.5em \@plus1ex \@minus.2ex}{-.5em}{\normalfont\normalsize\bfseries}}\makeatother

\definecolor{vis_red}{HTML}{EA4335}
\definecolor{vis_green}{HTML}{34A853}
\definecolor{vis_blue}{HTML}{4285F4}
\definecolor{vis_yellow}{HTML}{FBBC04}

\definecolor{light_red}{HTML}{FFD5D5}
\definecolor{light_green}{HTML}{E0EDDB}

\newcolumntype{x}{>{\columncolor{light_red}}c}
\newcolumntype{g}{>{\columncolor{light_green}}c}

\makeatletter
\DeclareRobustCommand\onedot{\futurelet\@let@token\@onedot}
\def\@onedot{\ifx\@let@token.\else.\null\fi\xspace}

\def\eg{\emph{e.g}\onedot} 
\def\ie{\emph{i.e}\onedot}

\newcommand{\figref}[1]{Figure~\ref{#1}}

\newcommand{\secref}[1]{Section~\ref{#1}}
\newcommand{\tabref}[1]{Table~\ref{#1}}

\definecolor{Gray}{gray}{0.9}
\definecolor{Grey}{gray}{0.7}
\definecolor{Half}{gray}{0.5}

\definecolor{lightred}{RGB}{255,213,213}
\definecolor{lightgreen}{RGB}{225,237,219}

\begin{document}

\title{Ego-Only: Egocentric Action Detection without Exocentric Transferring}

\author{Huiyu Wang\\
\and
Mitesh Kumar Singh\\
\and
Lorenzo Torresani \\
}
\author{
Huiyu Wang\textsuperscript{1}~~~~~~Mitesh Kumar Singh\textsuperscript{1}~~~~~~Lorenzo Torresani\textsuperscript{1}~~~~~~\\
\\
\textsuperscript{1}Meta AI\\
}

\maketitle

\begin{abstract}
We present Ego-Only, the \textbf{first} approach that enables state-of-the-art action detection on egocentric (first-person) videos without any form of exocentric (third-person) transferring. Despite the content and appearance gap separating the two domains, large-scale exocentric transferring has been the default choice for egocentric action detection. This is because prior works found that egocentric models are difficult to train from scratch and that transferring from exocentric representations leads to improved accuracy. However, in this paper, we revisit this common belief. Motivated by the large gap separating the two domains, we propose a strategy that enables effective training of egocentric models without exocentric transferring. Our Ego-Only approach is simple. It trains the video representation with a masked autoencoder finetuned for temporal segmentation. The learned features are then fed to an off-the-shelf temporal action localization method to detect actions. We find that this renders exocentric transferring unnecessary by showing remarkably strong results achieved by this simple Ego-Only approach on three established egocentric video datasets: Ego4D, EPIC-Kitchens-100, and Charades-Ego. On both action detection and action recognition, Ego-Only outperforms previous best exocentric transferring methods that use orders of magnitude more labels. Ego-Only sets new state-of-the-art results on these datasets and benchmarks without exocentric data.

\end{abstract}

\ificcvfinal\thispagestyle{empty}\fi

\section{Introduction}
\label{sec:intro}

\begin{figure}[!ht]
\setlength\tabcolsep{0.0pt}
\begin{tabular}{cc}
\small Ego4D Detection mAP & \small Charades-Ego Recognition mAP \\
\includegraphics[width=0.24\textwidth]{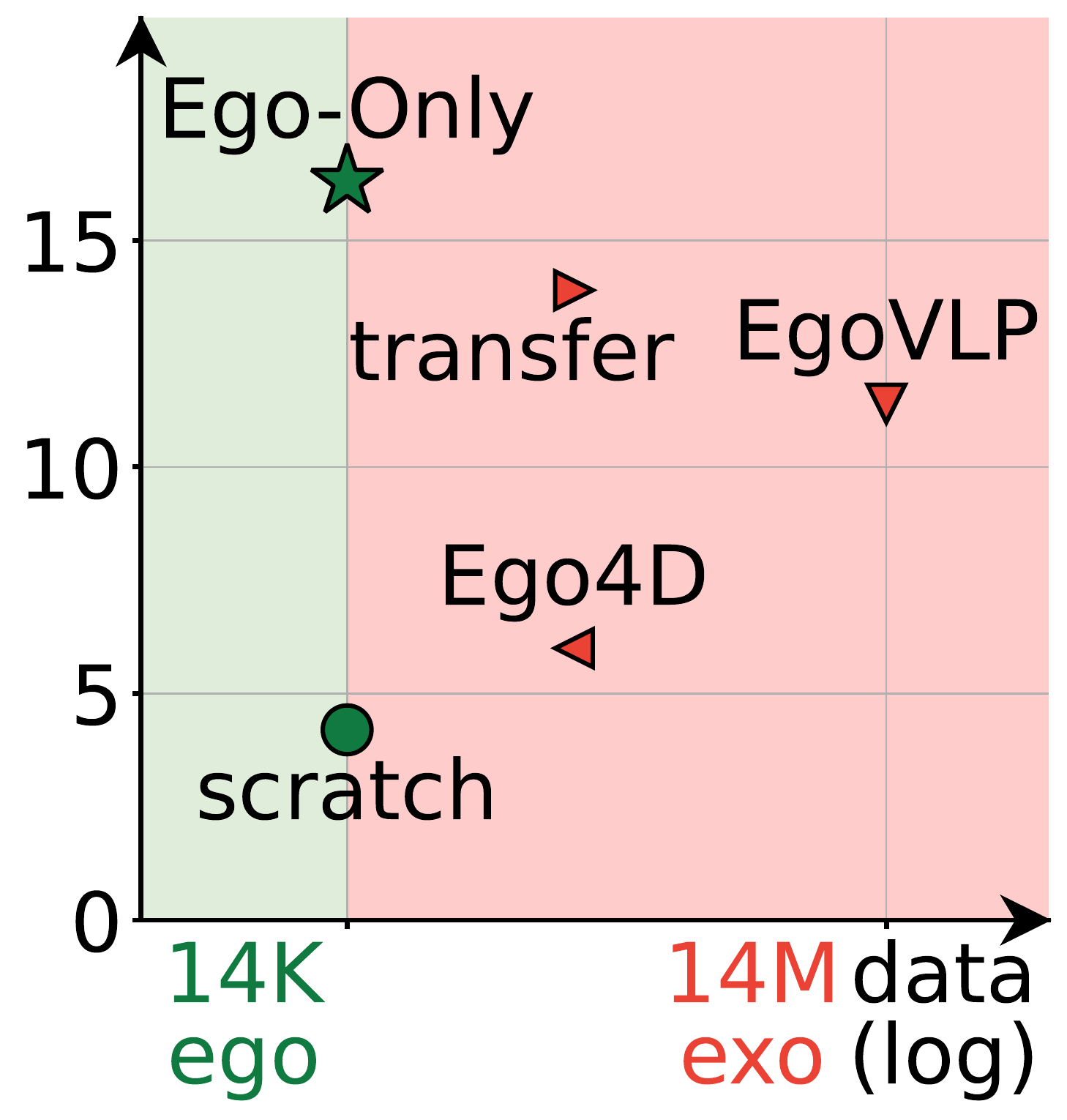} & \includegraphics[width=0.24\textwidth]{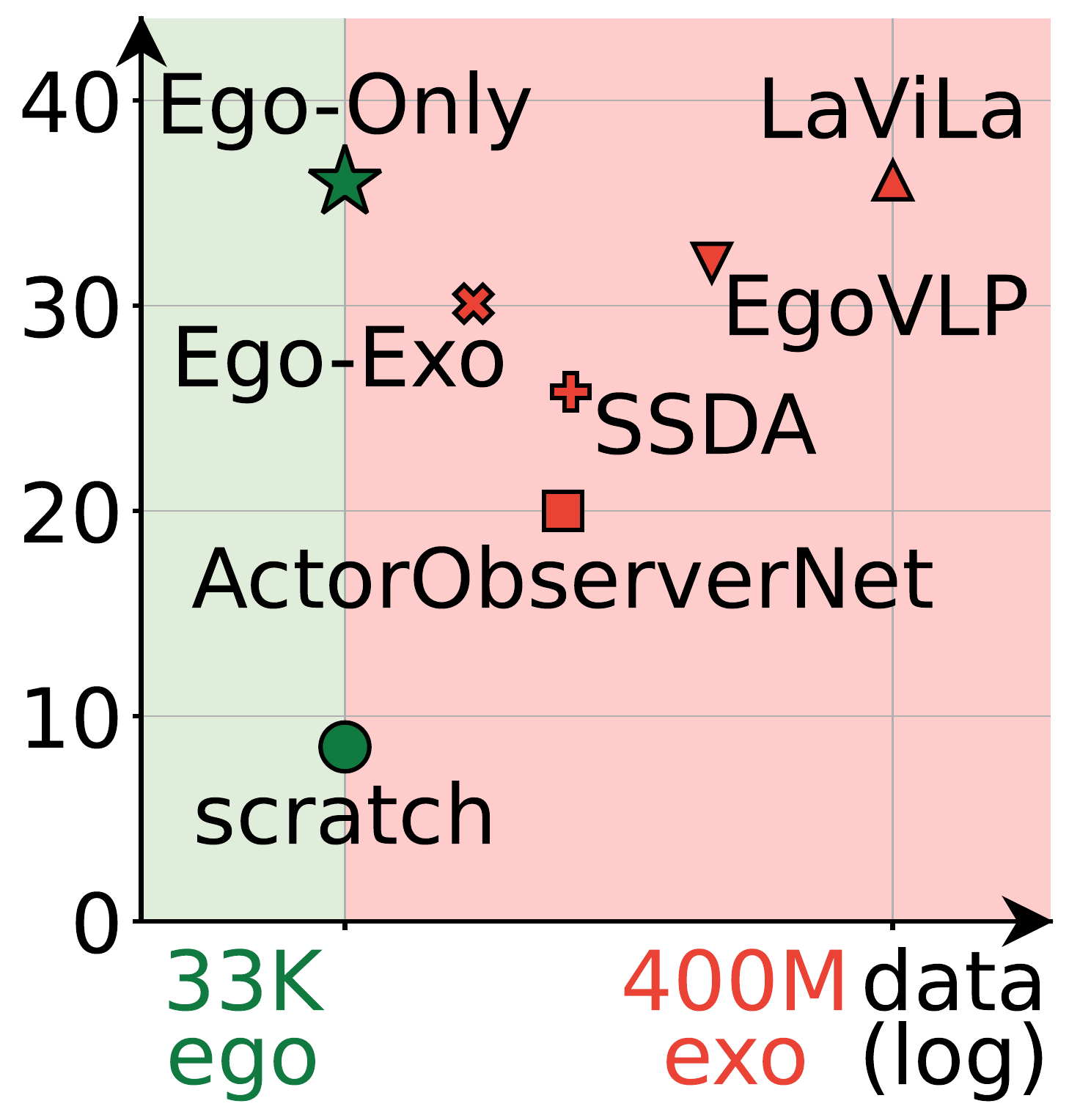} \\
\end{tabular}
\caption{Our Ego-Only approach achieves state-of-the-art results on Ego4D~\cite{ego4d} action detection and Charades-Ego~\cite{charades-ego} action recognition without any extra data or labels (\secref{sec:experiments}). Compared with exocentric transferring, Ego-Only uses orders of magnitude fewer labels, simplifies the pipeline,  and improves the results.}
\vspace{0ex}
\label{fig:teaser}
\end{figure}

\begin{figure}[!ht]
    \centering
    \setlength\tabcolsep{1.12345pt}
    \setlength\arrayrulewidth{0.08em}
    \begin{tabular}{gg|xx}
         & & & \\[-2.4ex]
        \multicolumn{2}{g|}{Egocentric Videos} & \multicolumn{2}{x}{Exocentric Videos} \\
        \multicolumn{2}{g|}{(length: 480 seconds)} & \multicolumn{2}{x}{(length: 10 seconds)} \\
        ~\includegraphics[height=0.075\textwidth]{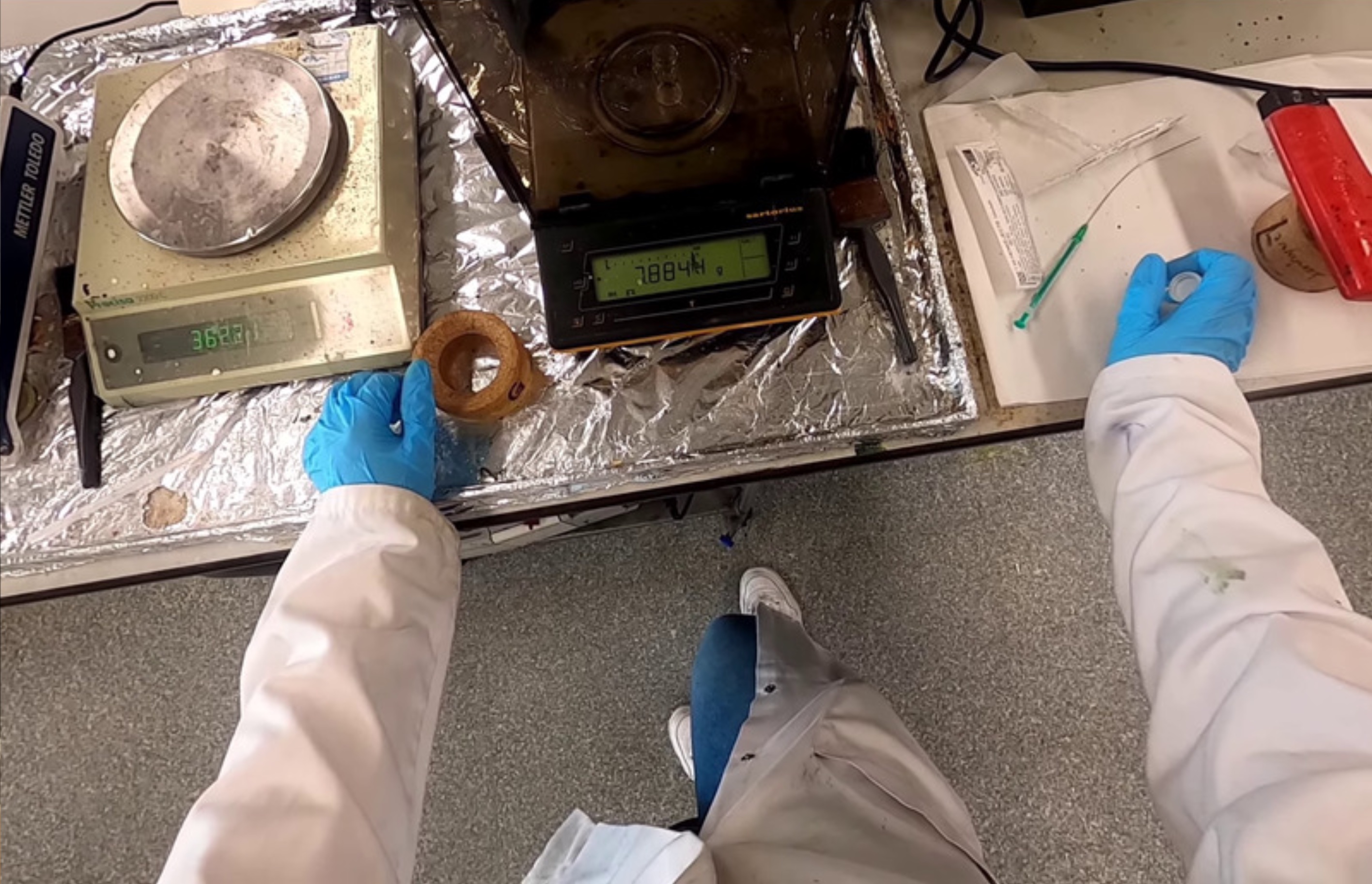} &
        \includegraphics[height=0.075\textwidth]{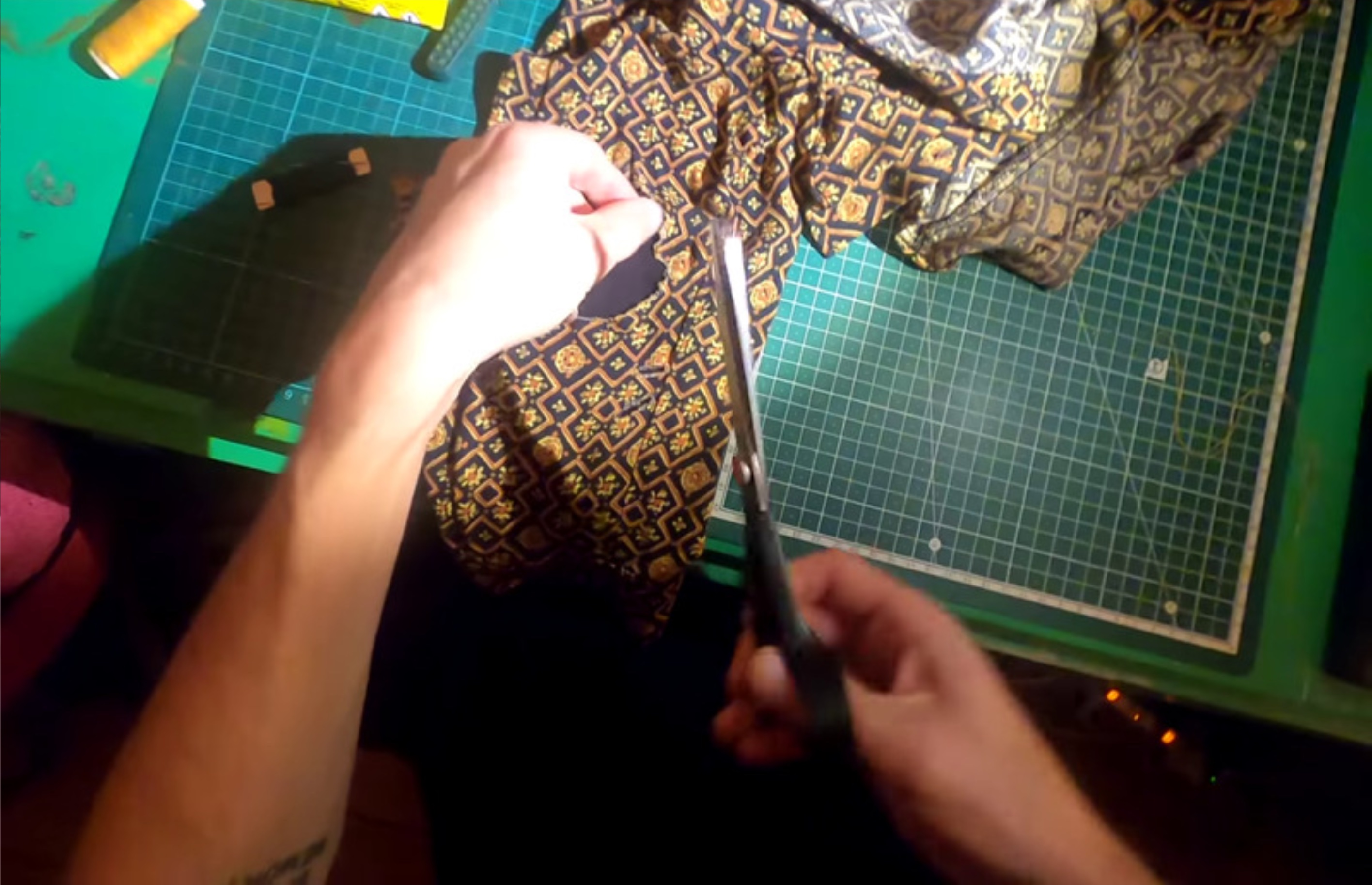}~ &
        ~\includegraphics[height=0.075\textwidth,width=0.10\textwidth]{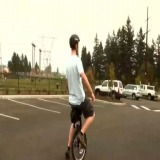} &
        \includegraphics[height=0.075\textwidth,width=0.10\textwidth]{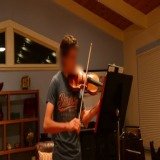}~ \\
        ~\includegraphics[height=0.075\textwidth]{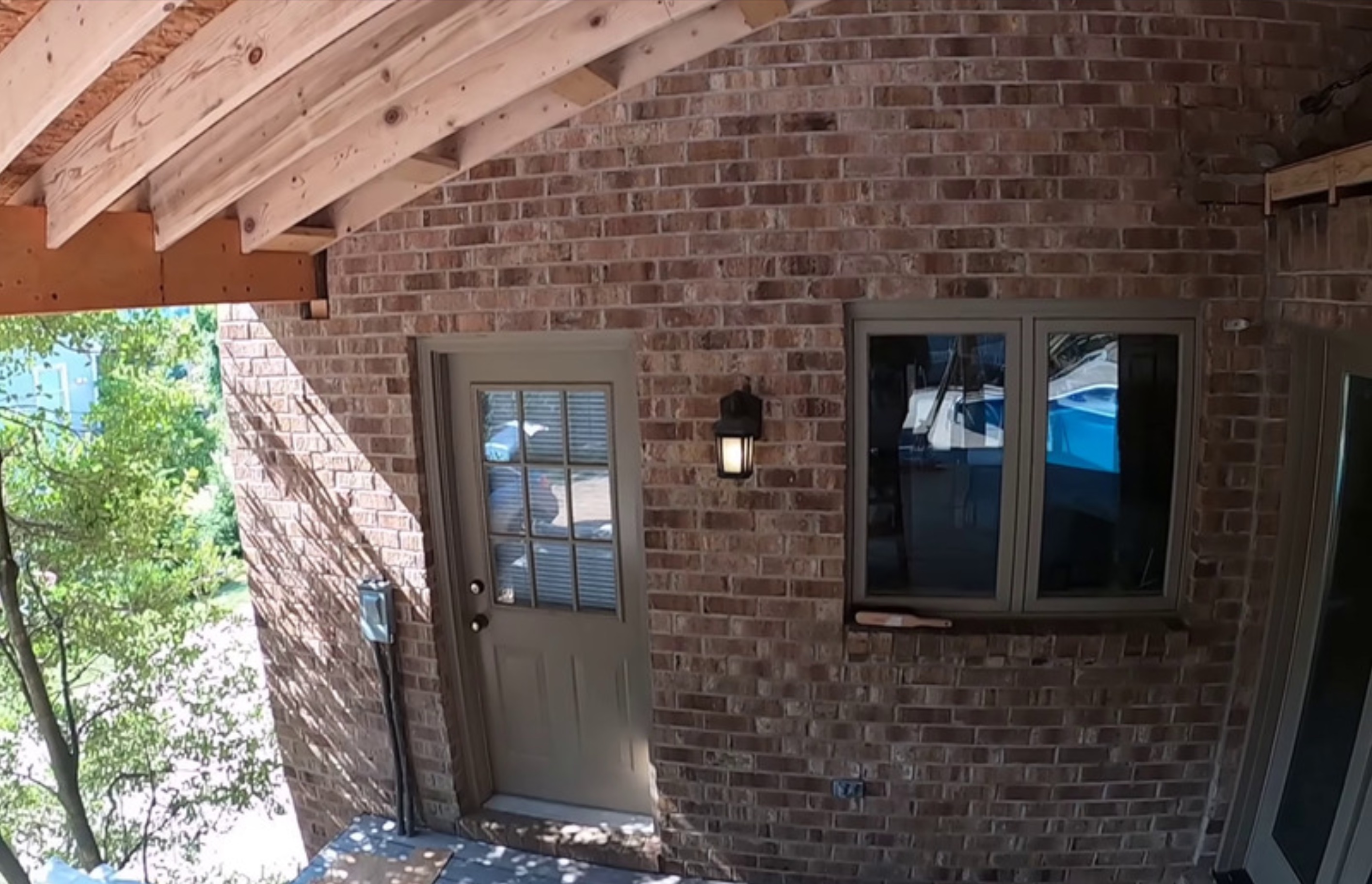} &
        \includegraphics[height=0.075\textwidth]{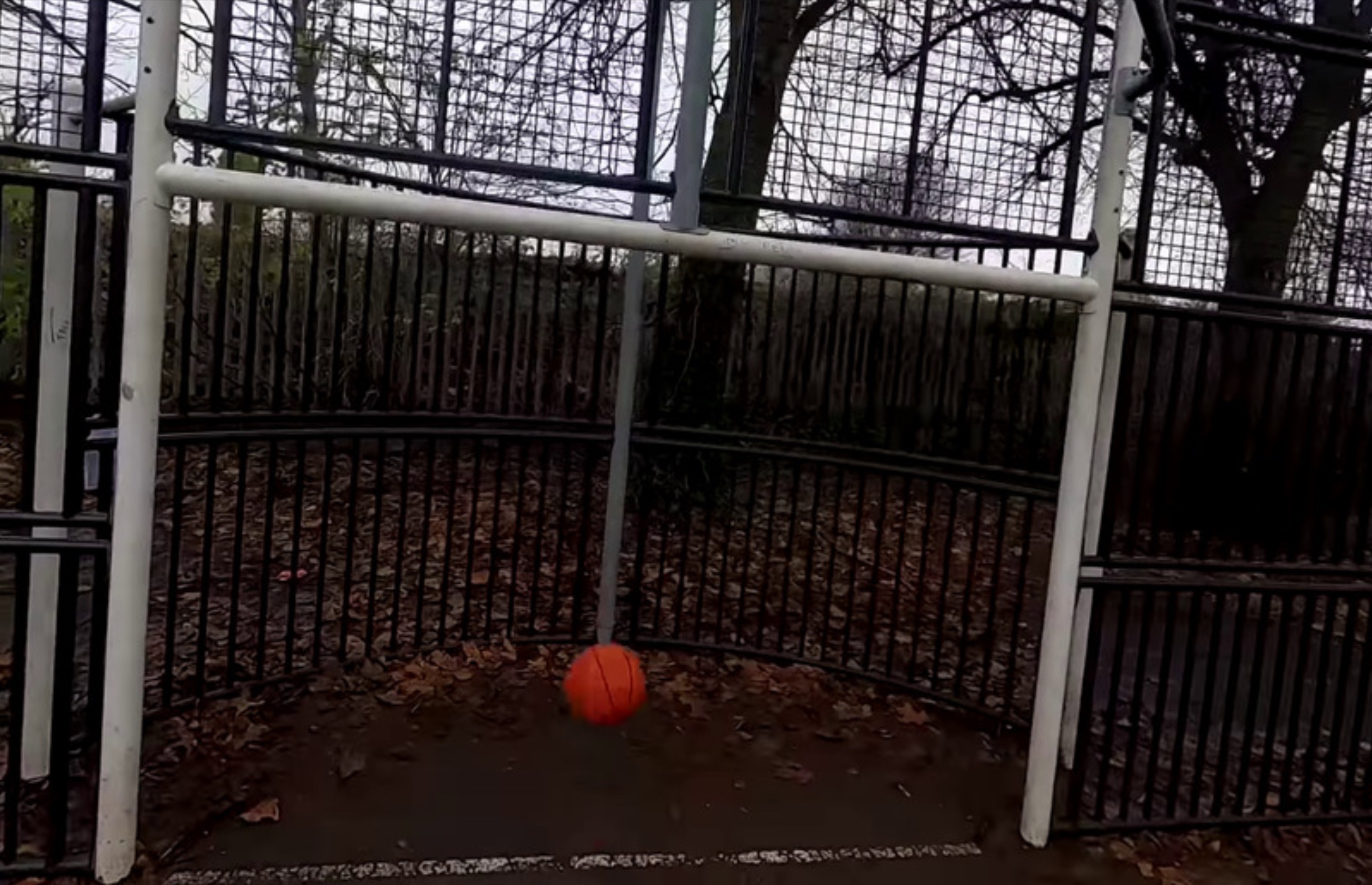}~ &
        ~\includegraphics[height=0.075\textwidth,width=0.10\textwidth]{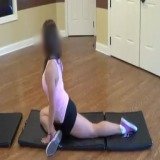} &
        \includegraphics[height=0.075\textwidth,width=0.10\textwidth]{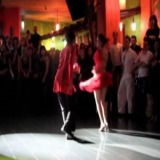}~ \\
    \end{tabular}%
    \vspace{1ex}
    \caption{Domain gap between egocentric videos (Ego4D~\cite{ego4d}) and exocentric videos (Kinetics-400~\cite{kay2017kinetics}). Exocentric videos are typically in the form of short trimmed clips, which show the actors as well as the contextual scene. Egocentric videos are dramatically longer, capture close-up object interactions but only the hands of the actor. These differences make it {\it challenging to transfer} models from exocentric action recognition to egocentric action detection.}
    \label{fig:example}
\end{figure}

In this paper we consider the problem of action detection from egocentric videos~\cite{ego4d,epic-kitchens-100,ego4d_data} captured by head-mounted devices. While action detection in third-person videos~\cite{activitynet,idrees2017thumos} has been the topic of extended and active research by the computer vision community, the formulation of this task in the first-person setting is underexplored. %

One major challenge of egocentric action detection is the lack of data, \ie insufficient amount of egocentric videos to train large-capacity models to competitive results. For example, existing methods such as Ego-Exo~\cite{ego-exo} and Charades-Ego~\cite{charades-ego}, attempted to train egocentric models from scratch using egocentric data only, but failed to obtain satisfactory results.
Therefore, current egocentric action detection methods rely on out-of-domain large-scale exocentric (third-person) videos~\cite{kay2017kinetics} or even images~\cite{imagenet}, under the assumption that the large-scale pretraining with proper transferring techniques can mitigate the negative effect of the domain gap between egocentric and exocentric videos. 
This hope is reinforced by the observation that deep neural networks exhibit invariance to object viewpoints~\cite{qiu2016unrealcv}, as evidenced by the effective transfers  from large-scale ImageNet pretraining to various still-image~\cite{lin2014microsoft,maskrcnn,maxdeeplab} and video understanding tasks~\cite{kay2017kinetics,timesformer,arnab2021vivit}.
Prior video approaches~\cite{ego-exo,charades-ego} also demonstrated empirical benefits of transferring from exocentric representations over simply learning egocentric representations from scratch.
As a result, this line of research focuses mainly on improving the transferring techniques that minimize the domain gap, or simply scaling exocentric data to a huge amount~\cite{yan2022multiview,omnivore}.

However, we argue that the dramatically different viewpoint of first-person videos poses challenges that may not be addressed simply by scaling exocentric data or designing better transferring techniques, as illustrated in \figref{fig:example}:
(1) No actor in view. In egocentric videos, the subject is behind the camera and is never visible, except for their hands. Conversely, third-person videos usually capture the actors as well as informative spatial context around them.
(2) Domain shift. Egocentric videos entail daily life activities such as cooking, playing, performing household chores, which are poorly represented in third-person datasets.
(3) Class granularity. First-person vision requires fine-grained recognition of actions within the same daily life category, such as ``wipe oil metallic item'', ``wipe kitchen counter'', ``wipe kitchen appliance'', and ``wipe other surface or object''~\cite{ego4d}.
(4) Object interaction. Egocentric videos capture a lot of human-object interactions as a result of the first-person viewpoint. The scales and views of the objects are dramatically different than in exocentric videos.
(5) Long-form. Egocentric videos are typically much longer than exocentric videos and thus require long-term reasoning of the human-object interactions rather than single frame classification.
(6) Long-tail. Real-world long-tail distribution is often observed in egocentric datasets, as they are uncurated and thus reflect the in-the-wild true distribution of activities, which is far from uniform.
(7) Localization. Egocentric action detection requires temporally sensitive representations which are difficult to obtain from third-person video classification on short and trimmed clips.

We argue that these challenges impede effective transfer from the exocentric to the egocentric domain and may actually cause detrimental biases when adapting third-person models to the first-person setting (as shown in \secref{sec:experiments}). Therefore, instead of following the common transferring assumption, we revisit the old good idea of training with in-domain egocentric data only, but this time in light of the development of recent data-efficient training methods, such as masked autoencoders~\cite{mae,videomae,feichtenhofer2022masked} as well as the scale growth of egocentric data collections (\eg, the recently introduced Ego4D dataset~\cite{ego4d}).

In this paper, we study the possibility of training with only egocentric video data by proposing a simple ``Ego-Only'' training approach. Specifically, Ego-Only consists of three training stages: (1) a masked autoencoder stage that bootstraps the backbone representation, (2) a simple finetuning stage that performs temporal semantic segmentation of egocentric actions, and (3) a final detection stage using an off-the-shelf temporal action detector, such as ActionFormer~\cite{actionformer}, without any modification. This approach enables us to train an egocentric action detector from random initialization without any exocentric videos or images.

Empirically, we evaluate Ego-Only on the three largest egocentric datasets, Ego4D~\cite{ego4d}, EPIC-Kitchens-100~\cite{epic-kitchens-100}, Charades-Ego~\cite{charades-ego}, and two tasks, action detection and action recognition. Surprisingly, Ego-Only outperforms all previous results based on exocentric transferring, setting new state-of-the-art results, obtained for the first time without additional data. Specifically, Ego-Only advances the state-of-the-art results on Ego4D Moments Queries detection (+6.5\% average mAP), EPIC-Kitchens-100 Action Detection (+5.5\% on verbs and +6.2\% on nouns), Charades-Ego action recognition (+3.1\% mAP), and EPIC-Kitchens-100 action recognition (+1.1\% top-1 accuracy on verbs).

In addition to the state-of-the-art comparison, we also noticed a few critical factors (as shown in \secref{sec:experiments}) for the effectiveness of an Ego-Only approach: (1) dramatic performance deterioration when skipping either MAE pretraining or temporal segmentation finetuning; (2) importance of MAE pretraining on egocentric (as opposed to exocentric) data to learn the in-domain distribution; (3) criticality of long-term modeling for good accuracy; (4) the sensitivity to amount of unsupervised data; (5) surprising lack of performance gains by joint ego-exo pretraining or finetuning.

In summary, our contributions are four-fold:
\begin{itemize}[nosep]
  \item We propose the first Ego-Only method that trains egocentric action representations effectively without any form of exocentric data or transferring.
  \item We demonstrate that exocentric transferring is \textit{not necessary} for state-of-the-art egocentric action detection. %
  \item Ego-Only advances state-of-the-art results on both action detection and action recognition, evaluated on three large-scale egocentric datasets.
  \item Our empirical evaluation reveals several critical factors for the effectiveness of an Ego-Only approach.
\end{itemize}

\section{Related Work}
\label{sec:related}

\paragraph{Action recognition} methods learn to classify actions in trimmed video clips. Recent action recognition models include convolutional neural networks~\cite{tran2015learning,carreira2017quo,wang2018temporal,tran2018closer,wang2018non,more-frames-2,slowfast,feichtenhofer2020x3d} and vision transformers~\cite{dosovitskiy2020image,timesformer,mvit,mvitv2,arnab2021vivit,liu2022video}. The learned action representations are often used as features for downstream tasks.

\paragraph{Temporal action localization} aims to detect action instances from long videos. Most methods~\cite{lin2017single,lin2019bmn,xu2020g,zhao2021video} detect actions using frozen video features from action recognition models. Recently, ActionFormer~\cite{actionformer} models long-sequence features with transformers. SegTAD~\cite{zhao2022segtad} detects actions via temporal segmentation. TALLFormer~\cite{cheng2022tallformer} trains the feature backbone end-to-end with the detector.

\paragraph{Self-supervised learning} aims to learn visual representation without human annotation. Traditional methods include hand-crafted pretext tasks~\cite{jigsaw,iterjigsaw,rotnet,pos} and contrastive learning~\cite{instdisc,moco,simclr,byol,simsiam,deepcluster,swav,dino,Wang_2022_CVPR}. Recently, masked autoencoders~\cite{beit,zhou2022ibot,mae,wei2022masked,feichtenhofer2022masked} have shown training efficiency~\cite{mae}, model scalability~\cite{mae}, data efficiency~\cite{videomae}, and effectiveness on videos~\cite{wei2022masked,videomae,feichtenhofer2022masked}.

\paragraph{Egocentric video} datasets~\cite{ego4d,damen2020epic,epic-kitchens-100,charades-ego-arxiv} have grown in size by orders of magnitude over the past few years, presenting new challenges~\cite{epic-kitchens-100} and opportunities\cite{ego4d}, such as egocentric action recognition~\cite{epic-kitchens-100,ego-exo} and detection~\cite{epic-kitchens-100}. Most egocentric action detection methods~\cite{ego4d,epic-kitchens-100,actionformer,egovlp} follow temporal action localization practices~\cite{zhao2021video,actionformer,lin2019bmn,xu2020g} and adopt exocentric pretrained checkpoints~\cite{carreira2017quo,slowfast,timesformer,frozenintime,arnab2021vivit}.

In this paper, we study the possibility of detecting egocentric actions without any form of exocentric transferring.

\section{Method}
\label{sec:method}

In \secref{sec:pipeline}, we provide an overview of our Ego-Only approach which enables egocentric action detection without relying on exocentric transferring. The proposed Ego-Only method consists of three training stages: a standard masked autoencoder (MAE) pretraining stage, an egocentric finetuning stage, which we present in \secref{sec:finetune}, and finally standard training of a temporal action detector.

\subsection{Ego-Only}
\label{sec:pipeline}

\begin{figure}[t]
    \centering
    \includegraphics[width=0.99\linewidth]{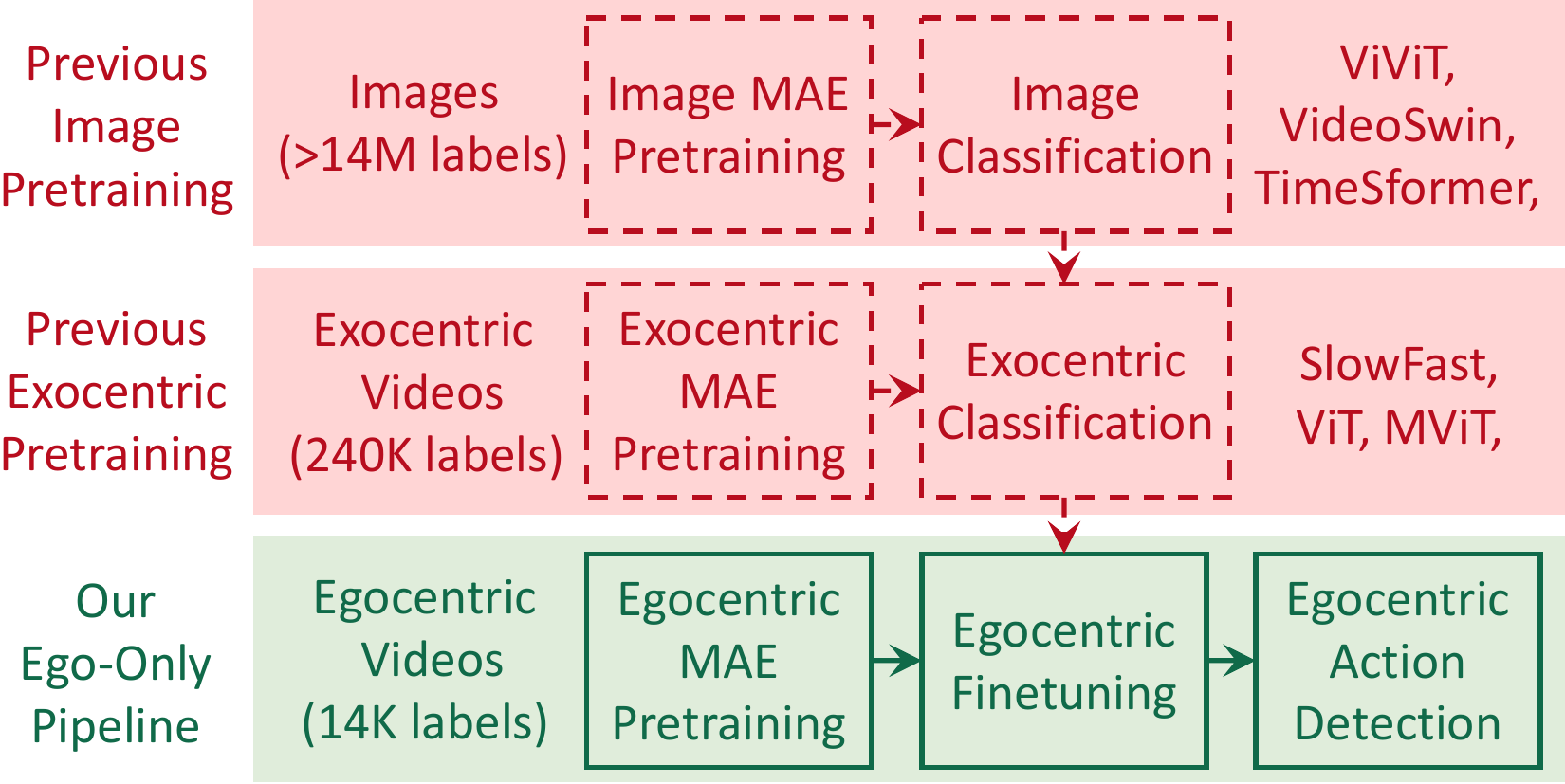}
    \caption{Our Ego-Only approach simplifies the previous pipeline by removing the dependence on pretrained exocentric checkpoints obtained with extra data, extra labels, and extra pretraining stages.}
    \vspace{-2ex}
    \label{fig:pipeline}
\end{figure}

\begin{figure*}[t]
    \centering
    \includegraphics[width=0.96\linewidth]{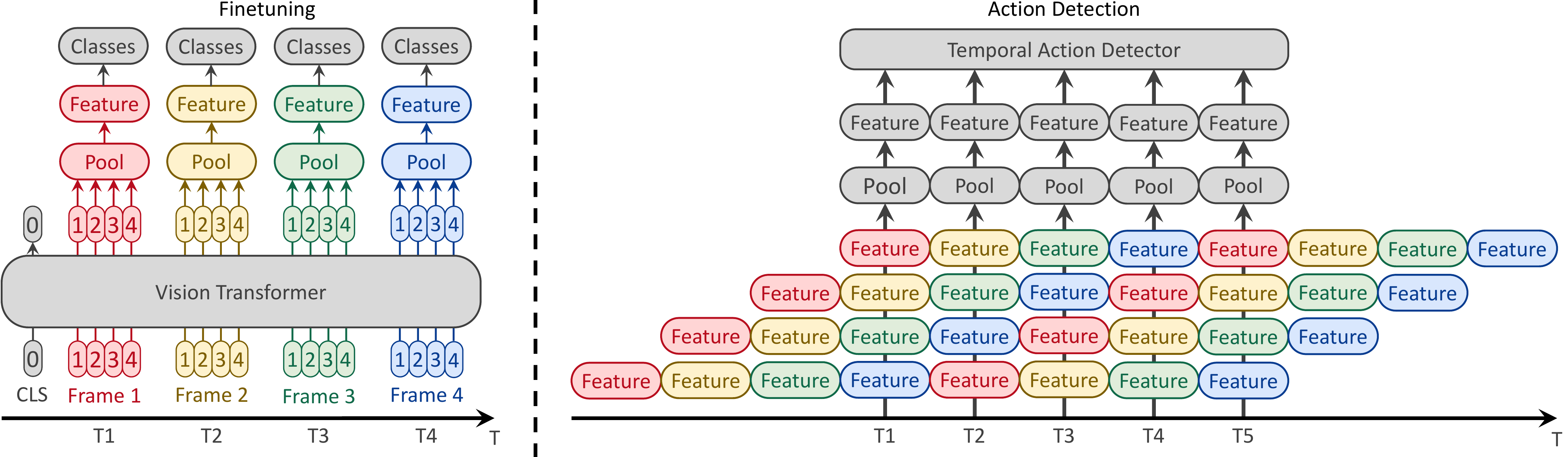}
    \caption{Ego-Only finetuning stage (left) and action detection stage (right). In the finetuning stage, the vision transformer is finetuned to predict action classes at each frame from spatially-pooled features (colors represent frame indices within a clip). In the detection stage, finetuned backbone features are frozen and extracted using a sliding window. Features at the same timestamp (\eg T1) but from different windows are average-pooled. On top of the long sequence of frozen features, a detector is then trained to temporally localize the actions.}
    \vspace{-1ex}
    \label{fig:seg}
\end{figure*}

There is an extensive literature about training object detectors~\cite{lin2014microsoft,maskrcnn} on images end-to-end from random initialization~\cite{he2019rethinking}. However, these approaches are difficult to adapt to egocentric action detection where both the videos and the actions are long-form. For example, Ego4D~\cite{ego4d} Moments clips are 8 minutes long, and around half of the actions are longer than 10 seconds which is the typical length of an exocentric video. In this case, end-to-end training of an action detector is impossible due to GPU memory limitations unless one reduces aggressively the model size, the spatial resolution, or the temporal sampling density, which would lead to degradation in performance.

This empirical challenge calls for a ``proxy'' objective that enables learning visual representations with a large model size, a high spatial resolution, and a high temporal sampling density. This surrogate objective is usually realized by pretraining on short exocentric videos. However, as discussed in \secref{sec:intro}, the learned representation may not transfer effectively. Instead, in our Ego-Only approach, we approximate the temporal action detection task by performing temporal semantic segmentation that predicts action labels at each frame. Note that this approximation is not exact because we truncate long-form videos into clips, throwing away the action context outside the sampled clip. Such approximation leads to a trade-off between the action context and the temporal sampling density, ablated in \secref{sec:ablation}.

This simple surrogate objective allows us to train visual representations from random initialization towards temporal action detection. However, we empirically find that the learned representation generalizes poorly even with strong augmentation and regularization. In order to further improve generalization, we introduce an additional MAE pretraining stage which has been shown to yield strong generalization in the low-data regime~\cite{videomae}. This additional pretraining improves generalization as shown in \tabref{tab:pretraining}.

Putting these pieces together, \figref{fig:pipeline} summarizes our complete Ego-Only method that includes the initial MAE pretraining, the egocentric finetuning task as an approximation of action detection, and the final temporal action detector that incorporates full context of the whole long-form video. This approach differs from existing methods in the absence of an exocentric pretraining stage that requires large-scale annotated exocentric videos or images. For example, most prior approaches pretrain egocentric models on Kinetics-400 (K400) with 240K annotated videos, while our Ego-Only method uses merely 14K annotated action segments on Ego4D and achieves better results (\tabref{tab:pretraining}).

Next, we describe in more detail the initial MAE pretraining stage and the final action detection stage that are both adopted from existing literature without any modification. Note that this paper aims to revisit the value of exocentric transferring and does so by proposing an ego-only meta algorithm that is intentionally kept as simple as possible.

\paragraph{Masked Autoencoder.}

Our method applies the original MAE~\cite{mae} and video MAE~\cite{feichtenhofer2022masked} algorithms. Specifically, we consider the vanilla vision transformers~\cite{dosovitskiy2020image,feichtenhofer2022masked}, ViT-B and ViT-L, as our architectures, due to the native support by MAE. We do not consider convolutional architectures~\cite{slowfast} or hierarchical transformers~\cite{swin,liu2022video,mvit,mvitv2} that require adaptation of the MAE algorithm. Since videos are highly redundant, we use a very high masking ratio (90\%) with a random masking strategy and all of the pretraining recipes as suggested in video MAE~\cite{feichtenhofer2022masked}. The only adaptation we make is to sample each video with a probability proportional to its temporal length, because of the long-form property of egocentric videos. This ensures equal sampling probability for any possible clip in the dataset.

\paragraph{Action Detector.}

After the egocentric finetuning stage (\secref{sec:finetune}) that trains the backbone representation towards action detection, we apply an existing temporal action localization algorithm to detect the actions. Specifically, given the finetuned video backbone, features are extracted from the frozen model with sliding windows, following standard practice in temporal action localization~\cite{actionformer,zhao2021video}. Then, the action detector is trained on top of a long sequence of frozen video features to produce temporal segments as outputs. There is a potential risk of overfitting since our finetuning stage and action detection stage are trained on the same training set, but empirically 
we do not find this to be a significant issue in practice, probably because the detector takes as input a long-form video instead of a clip and the detector loss differs from simply segmentation. For better performance, we choose ActionFormer~\cite{actionformer} as our default detector as it has demonstrated good accuracy on temporal action localization benchmarks. As we work on egocentric videos, we adopt the ActionFormer architecture previously proposed for EPIC-Kitchens-100~\cite{epic-kitchens-100}.

\subsection{Finetuning via Temporal Segmentation}
\label{sec:finetune}

Inspired by TSN~\cite{wang2018temporal} and SegTAD~\cite{zhao2022segtad} that detect actions via temporal semantic segmentation, we finetune our backbone features from MAE pretraining by predicting class labels for each frame, as illustrated in \figref{fig:seg}~(left). This is akin to the task of image semantic segmentation~\cite{deeplabv12015,chen2018deeplabv2,chen2017deeplabv3,deeplabv3plus2018} which predicts class labels for each pixel. Formally, given an input video clip with a certain temporal span, a temporal segmentation model predicts output logits $L \in \mathbb{R}^{T\times C}$ where $T$ denotes the temporal dimension of the logits and $C$ is the total number of action classes.

We follow a few principles in defining this simple finetuning objective: (1) A video clip of a certain temporal span is taken as the input instead of the full long-form video. This temporal approximation enables us to train large-scale models within the given GPU memory limit. (2) We employ a fixed temporal span which is consistent with both MAE pretraining and detection feature extraction. This removes potential domain gaps when models are trained and inferred with different temporal spans. (3) The temporal segmentation objective trains models to distinguish frames of different classes within one video clip, especially when a long temporal span is adopted. (4) We train with clips uniformly sampled over the dataset, making full use of all positive and negative samples in the dataset.

Note that our segmentation stage differs from TSN~\cite{wang2018temporal} and SegTAD~\cite{zhao2022segtad} mainly in the goal which is to finetune the backbone representation instead of to detect actions directly from the output scores. In order to address unique challenges (\secref{sec:intro}) in egocentric videos, we also adopt critical techniques addressing loss and imbalance issues.

Next, we discuss the loss function that we choose to finetune the backbone, how we address the egocentric imbalance challenges, and how backbone features are extracted for the subsequent action detection stage.

\paragraph{Loss function.}

Egocentric videos usually contain overlapping actions of different classes. For example, a person could be taking a photo while speaking on the phone. This makes the finetuning stage a multi-label classification task. Therefore, we employ a loss function independent for each action class, \ie the activation of one class does not suppress another. Specifically, we adopt per-frame binary cross-entropy (BCE) as the loss function on the logits, instead of cross-entropy which suppresses non-maximum classes.

\paragraph{Imbalance challenges.}

The long-tail imbalance in egocentric videos (\secref{sec:intro}) poses a major challenge to our finetuning stage, due to the less curated nature and the long-form property of egocentric videos. Specifically, there are usually (1) imbalanced numbers of videos across action classes, (2) imbalanced action lengths within one class, and (3) imbalanced numbers of foreground frames vs background frames within one class. Inspired by the literature of one-stage object detection, we mitigate the imbalance issue by adopting focal loss~\cite{lin2017focal} in the BCE loss and biasing the logits towards background at initialization. We also reweigh each action instance by the inverse of the action length, leading to a balanced loss for each instance.

\paragraph{Feature extraction.}

Once our video backbone is finetuned on sampled clips, features are extracted using a sliding window on both the training set and validation set for training the detector on long-form videos and validating the approach. According to temporal action localization literature~\cite{actionformer,zhao2021video}, clip features are average-pooled spatiotemporally following the exocentric classification practice~\cite{carreira2017quo,slowfast}. However, in our temporal segmentation case on long-form videos, our spatially-pooled features are trained to be temporally different within a video clip, encoding their own local context. Therefore, as illustrated in \figref{fig:seg}~(right), given the sliding windows of features, we average-pool features at the same wall-clock timestamp from all sliding windows. This enables the usage of a long temporal span, such as 64 seconds (\figref{fig:span}), by extracting temporally variable features from a window. %

\section{Experiments}
\label{sec:experiments}

We evaluate our Ego-Only approach by reporting main results on the two largest egocentric video datasets, Ego4D~\cite{ego4d} and EPIC-Kitchens-100~\cite{epic-kitchens-100}, measured by average mAP at tIoU \{0.1, 0.2, 0.3, 0.4, 0.5\} on the val set (\secref{sec:detection}). Then, we study the application to egocentric action recognition and report video-level mAP on Charades-Ego~\cite{charades-ego} and top-1 accuracy on EPIC-Kitchens-100~\cite{epic-kitchens-100} (\secref{sec:recognition}). Finally, we carefully ablate the effect of each design choice in \secref{sec:ablation}. %

\subsection{Main Results on Action Detection}
\label{sec:detection}

\begin{table*}[t]
\begin{center}
\setlength\tabcolsep{3.345pt}
\begin{tabular}{l|ccc|ccccc|cc}
\toprule
method & backbone & extra data & extra labels & 0.1 & 0.2 & 0.3 & 0.4 & 0.5 & \small avg & \# labels \\
\midrule
Ego4D~\cite{ego4d} & SlowFast~\cite{slowfast} & Kinetics-400~\cite{kay2017kinetics} & 240K & 9.10 & 7.16 & 5.76 & 4.62 & 3.41 & 6.03 & 254K\\
EgoVLP~\cite{egovlp} & Frozen~\cite{frozenintime} & \small IN-21K~\cite{imagenet} + EgoClip~\cite{egovlp} & 18M & 16.63 & - & 11.45 & - & 6.57 & 11.39 & 18M \\
\midrule
Ego-Only & ViT-B & - & - & 22.5 & 19.3 & 16.0 & 13.1 & 10.6 & 16.3 & \bf 14K \\
Ego-Only & ViT-L & - & - & 24.6 & 20.8 & 17.7 & 14.9 & 11.7 & \bf 17.9 & \bf 14K \\
\bottomrule
\end{tabular}
\end{center}
\caption{Ego4D action detection on MQ val set.
}
\vspace{-0.5ex}
\label{tab:ego4d}
\end{table*}

\paragraph{Ego4D.}
We compare our results on the Ego4D~\cite{ego4d} MQ val set with state-of-the-art methods in \tabref{tab:ego4d}, using ViT-B and ViT-L. We notice that our Ego-Only performs significantly better than previous state-of-the-art but without any extra exocentric data or labels needed.
Specifically, with ViT-B as the backbone, Ego-Only achieves an average mAP of 16.3\%, producing a relative improvement of 170\% over the Ego4D paper baseline~\cite{ego4d} that pretrains on Kinetics-400~\cite{kay2017kinetics} with 18$\times$ annotated clips.
This strong result even outperforms EgoVLP which has seen 4M language-narrated video clips from Ego4D (\ie in-domain) and 14M images from IN-21K~\cite{imagenet}.
Finally, scaling Ego-Only to ViT-L backbone yields an mAP of 17.9\%, setting a new state-of-the-art on this benchmark without any extra data or labels.

\paragraph{EPIC-Kitchens-100.}

\begin{table*}[t]
\begin{center}
\setlength\tabcolsep{3.12345pt}
\begin{tabular}{l|ccc|ccccc|c|ccccc|c|c}
\toprule
\multirow{2}{*}{method} & \multirow{2}{*}{backbone} & extra & extra & \multicolumn{6}{c|}{verb} & \multicolumn{6}{c|}{noun} & \# labels \\
 & & data & labels & 0.1 & 0.2 & 0.3 & 0.4 & 0.5 & avg & 0.1 & 0.2 & 0.3 & 0.4 & 0.5 & avg & seen \\
\midrule[0.08em]
BMN~\cite{lin2019bmn,epic-kitchens-100} & SlowFast & K400 & 240K & 10.8 & 9.8 & 8.4 & 7.1 & 5.6 & 8.4 & 10.3 & 8.3 & 6.2 & 4.5 & 3.4 & 6.5 & 307K \\
G-TAD~\cite{xu2020g} & SlowFast & K400 & 240K & 12.1 & 11.0 & 9.4 & 8.1 & 6.5 & 9.4 & 11.0 & 10.0 & 8.6 & 7.0 & 5.4 & 8.4 & 307K \\
ActionFormer & SlowFast & K400 & 240K & 26.6 & 25.4 & 24.2 & 22.3 & 19.1 & 23.5 & 25.2 & 24.1 & 22.7 & 20.5 & 17.0 & 21.9 & 307K \\
\midrule
Ego-Only & ViT-B & - & - & 31.1 & 30.4 & 28.9 & 26.6 & 23.4 & 28.1 & 30.0 & 29.2 & 27.8 & 25.1 & 20.7 & 26.5 & \bf 67K \\
Ego-Only & ViT-L & - & - & 32.0 & 31.5 & 20.0 & 27.4 & 24.0 & \bf 29.0 & 31.5 & 30.8 & 29.2 & 26.5 & 22.5 & \bf 28.1 & \bf 67K \\

\bottomrule
\end{tabular}
\end{center}
\caption{EPIC-Kitchens-100 Action Detection val set.
}
\vspace{-1ex}
\label{tab:epic}
\end{table*}

Following the Ego4D exploration, we validate our Ego-Only approach on the EPIC-Kitchens-100~\cite{epic-kitchens-100} Action Detection benchmark. We can see from \tabref{tab:epic} that Ego-Only achieves much better results compared with exocentric transferring.
Specifically, compared with previous state-of-the-art methods that adopt Kinetics~\cite{kay2017kinetics} SlowFast~\cite{slowfast} features finetuned on EPIC-Kitchens-100 Action Recognition, our Ego-Only with a ViT-B backbone already performs 4.6\% better on both verbs and nouns. Scaled to a ViT-L backbone, Ego-Only improves further and sets a new state-of-the-art result of 29.0\% mAP on verbs and 28.1\% mAP on nouns. By analyzing our results using DETAD~\cite{alwassel_2018_detad} in \secref{sec:analysis}, we find that Ego-Only significantly reduces false positives on backgrounds, compared with exocentric transferring, probably because Kinetics contains mostly trimmed videos with foreground actions only. This validates the benefit of Ego-Only.
\subsection{Application to Action Recognition}
\label{sec:recognition}

Besides egocentric action detection, we further evaluate our Ego-Only approach on the task of action recognition on Charades-Ego~\cite{charades-ego} and EPIC-Kitchens-100~\cite{epic-kitchens-100}. This is simply achieved by skipping our last action detector stage and averaging the temporal semantic segmentation model output scores after the sigmoid activation in the BCE loss. Results on action recognition allow us to compare Ego-Only with a wider range of state-of-the-art methods.

\paragraph{Charades-Ego.}

\begin{table}[t]
\begin{center}
\setlength\tabcolsep{3.02345pt}
\begin{tabular}{l|cc|cc}
\toprule
method & backbone & params & mAP & \# labels \\
\midrule
\small ActorObserver~\cite{charades-ego} & ResNet-152 & 60M & 20.0 & 1.4M \\ %
SSDA~\cite{choi2020unsupervised} & I3D & 12M & 25.8 & 1.6M \\ %
Ego-Exo~\cite{ego-exo} & \small SlowFast-R101 & 75M & 30.1 & 0.3M \\  %
EgoVLP~\cite{egovlp} & TSF-B & 178M & 32.1 & 18M \\
\midrule
LaViLa~\cite{zhao2022lavila} & TSF-B & 178M & 33.7 & 404M \\
Ego-Only & ViT-B & 87M & 33.3 & \bf 33K \\
\midrule
LaViLa~\cite{zhao2022lavila} & TSF-L & 528M & 36.1 & 404M \\
Ego-Only & ViT-L & 304M & 36.0 & \bf 33K \\
Ego-Only\textsuperscript{\textdagger} & ViT-L & 304M & \bf 39.2 & \bf 67K \\
\bottomrule
\end{tabular}
\end{center}
\caption{Charades-Ego recognition. \textsuperscript{\textdagger}with full Charades-Ego data.
}
\vspace{-0.5ex}
\label{tab:charadesego}
\end{table}

In \tabref{tab:charadesego}, we report recognition results on Charades-Ego~\cite{charades-ego} by finetuning the existing Ego4D MAE checkpoints on Charades-Ego, without exploiting any ego-exo supervision or correspondence. Remarkably, Ego-Only with a ViT-B backbone already significantly outperforms state-of-the-art methods that exploit ego-exo alignment (ActorObserverNet~\cite{charades-ego}), or semi-supervised domain adaptation (SSDA~\cite{choi2020unsupervised}), or ego-exo distillation (Ego-Exo~\cite{ego-exo}), or egocentric video-language pretraining (EgoVLP~\cite{egovlp}). Furthermore, we compare LaViLa that uses CLIP initialization with 400M text-image pairs, 4M Ego4D narration-clip pairs, as well as the large language model GPT-2 XL. Our Ego-Only trained on only the egocentric subset of Charades-Ego, matches this result with merely 33K labels (around 0.01\% of 404M) and a smaller ViT-L backbone. Finally, when we augment Ego-Only with the exocentric subset of Charades-Ego, we observe a significant gain of 3.1\% absolute points over the LaViLa state-of-the-art.

\paragraph{EPIC-Kitchens-100.}

\begin{table}[t]
\begin{center}
\setlength\tabcolsep{2.12345pt}
\begin{tabular}{ll|cc}
\toprule
method & variant & verb & noun \\
\midrule
IPL~\cite{wang2021interactive} & I3D, K400 & 68.6 & 51.2 \\
ViViT~\cite{arnab2021vivit} & ViViT-L/16x2, IN-21k+K400 & 66.4 & 56.8 \\
MoViNet~\cite{kondratyuk2021movinets} & MoViNet-A6, 120 frames & 72.2 & 57.3 \\
MTV~\cite{yan2022multiview} & MTV-B, WTS-60M, 280p & 69.9 & 63.9 \\
MTCN~\cite{kazakos2021little} & \scriptsize MFormer-HR, IN-21k+K400+VGG-Sound & 70.7 & 62.1 \\
Omnivore~\cite{omnivore} & \footnotesize Swin-B, IN21k+IN-1k+K400+SUN & 69.5 & 61.7 \\
MeMViT~\cite{memvit} & \small MeMViT, 32$\times$3, K600, 105.6 sec & 71.4 & 60.3 \\
LaViLa~\cite{zhao2022lavila} & \small TSF-L, WebImageText+Ego4D & 72.0 & 62.9 \\
\midrule
Ego-Only & ViT-L, 32 frames, 3.2 sec & \bf 73.3 & 59.4 \\
\bottomrule
\end{tabular}
\end{center}
\caption{EPIC-Kitchens-100 action recognition top-1 accuracy.
}
\vspace{-0.5ex}
\label{tab:epic_rec}
\end{table}

In \tabref{tab:epic_rec}, we report action recognition top-1 accuracies on EPIC-Kitchens-100~\cite{epic-kitchens-100} by evaluating the EPIC-Kitchens-100 temporal segmentation model from \secref{sec:detection}. We compare Ego-Only with state-of-the-art methods exploiting large-scale image data (ViViT~\cite{arnab2021vivit}), or web-scale text-image pairs (MTV~\cite{yan2022multiview}, LaViLa~\cite{zhao2022lavila}), or multimodal audio (MTCN~\cite{kazakos2021little}) depth (Omnivore~\cite{omnivore}) supervision, or 32$\times$ temporal support (MeMViT~\cite{memvit}). In contrast, our Ego-Only using the 495 videos in EPIC-Kitchens-100 as the only source of supervision achieves the state-of-the-art results of 73.3\% on verb classification, outperforming the existing best result by 1.1\%. This validates the effectiveness of Ego-Only in capturing hand-object interactions from egocentric videos.
\subsection{Ablation Study}
\label{sec:ablation}

In order to analyze our Ego-Only approach, we compare Ego-Only with common exocentric transferring solutions and ablate the importance of each stage in Ego-Only. We also scale the amount of data consumed, the model sizes, as well as the number of pretraining epochs. We perform all ablation studies on egocentric action detection benchmarks.

\begin{table}[t]
\begin{center}
\setlength\tabcolsep{4.12345pt}
\small
\begin{tabular}{c|ccc|cc}
\toprule
\multirow{2}{*}{method} & self-sup. & sup. & sup. & Ego4D & \# labels \\
 & MAE & exo & ego & mAP & seen \\
\midrule
exo-sup & - & K400 & Ego4D & 13.9 & 254K (18$\times$)\\
\rowcolor{Gray}
ours & Ego4D & - & Ego4D & \bf 16.3 & \bf 14K (1$\times$)\\
\midrule
\color{Grey} scratch & \color{Grey} - & \color{Grey} - & \color{Grey} Ego4D & \color{Grey} 4.2 & \color{Grey} 14K (1$\times$)\\
\color{Grey} exo-MAE & \color{Grey} K400 & \color{Grey} - & \color{Grey} Ego4D & \color{Grey} 13.4 & \color{Grey} 14K (1$\times$)\\
\color{Grey} exo-FT & \color{Grey} K400 & \color{Grey} K400 & \color{Grey} Ego4D & \color{Grey} 16.2 & \color{Grey} 254K (18$\times$)\\
\bottomrule
\end{tabular}
\end{center}
\caption{Varying the pretraining stage. Ego-Only outperforms exocentric transferring with much fewer labels (14K vs. 240K+14K).}
\vspace{-0.5ex}
\label{tab:pretraining}
\end{table}

\paragraph{Varying the pretraining stage.}

\tabref{tab:pretraining} reports our results with different pretraining stages. Compared with the common exocentric supervised baseline of 13.9\% mAP, our Ego-Only with exactly the same backbone, the same finetuning, and the same detector, achieves the performance of 16.3\% (+2.4\%) mAP by using egocentric data only and with merely 14K labels, instead of 240K labels used in the exocentric transferring method.

Next, we consider skipping the MAE pretraining and train from scratch the model via temporal segmentation on Ego4D. However, our best model learned from scratch only reaches the mAP of 4.2\% (vs. 16.3\% with MAE pretraining in Ego-Only), due to the limited number of labels available on Ego4D, only 14K. This is smaller than the number of labels in MNIST~\cite{mnist} or CIFAR~\cite{cifar} but the task of egocentric action detection is significantly more challenging.

In addition to the model trained from scratch, we also compare with self-supervised MAE pretraining on Kinetics-400. When this checkpoint is finetuned, it achieves 13.4\% mAP which is 2.9\% worse than the counterpart pretrained on Ego4D. This gap is reasonable since the model is pretrained on out-of-domain data but does not benefit from the large-scale exocentric labels. Once the extra labels are used, Kinetics finetuning yields performance on-par with our much simpler Ego-Only approach.

\begin{table}[t]
\begin{center}
\small
\begin{tabular}{c|ccc|c}
\toprule
\multirow{2}{*}{method} & self-sup. & sup. & sup. & Ego4D \\
 & MAE & exo & ego & mAP \\
\midrule
exo-MAE & K400 & - & - & 6.7 \\
ego-MAE & Ego4D & - & - & 7.8 \\
exo-FT & K400 & K400 & - & 13.5 \\
\rowcolor{Gray}
ours & Ego4D & - & Ego4D & 16.3 \\
\bottomrule
\end{tabular}
\end{center}
\caption{Varying the finetuning stage.}
\vspace{-0.5ex}
\label{tab:finetuning}
\end{table}

\paragraph{Varying the finetuning stage.}

After varying the pretraining stage, we study the importance of finetuning. For this purpose, we extract features from pretrained models, without any form of finetuning on egocentric data. Contrary to the strong linear probing results of MAE on ImageNet-1K\cite{imagenet}, we observe that frozen MAE features perform poorly on egocentric action detection, leading to an absolute drop of 8.5\% points in average mAP. Kinetics-400 MAE features perform even worse (as expected), but finetuning on Kinetics with 240K labels is helpful, achieving a 13.5\% mAP which is 2.8\% worse than Ego-Only. We also try concatenating frozen MAE features from multiple blocks, inspired by DINO\cite{dino}, but only observe a marginal gain in \secref{sec:concat}. %

\begin{figure}[t]
    \centering
    \includegraphics[width=0.43\textwidth]{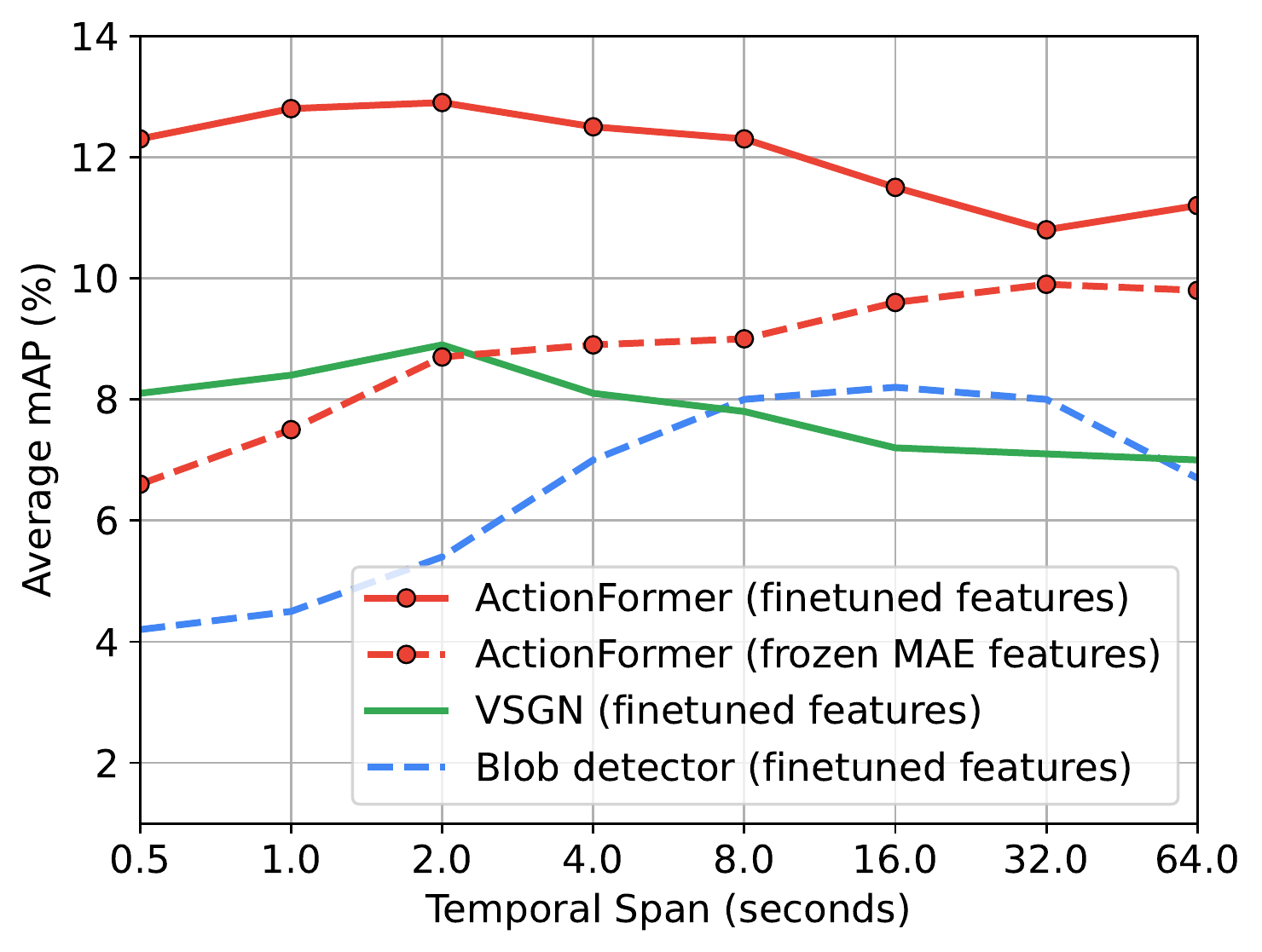}
    \caption{Varying detectors and temporal spans. The blob detector performs surprisingly well and perfers a long temporal span, while ActionFormer and VSGN prefer short spans due to their transformer or graph neural network based architectures.}
    \vspace{-0.5ex}
    \label{fig:span}
\end{figure}

\paragraph{Detectors and temporal spans.}

Next, we compare temporal action detector choices in Ego-Only and vary the temporal span at the same time. As we use a consistent temporal span for the whole pipeline, including MAE, finetuning, and feature extraction (\secref{sec:finetune}), we pretrain MAE with each temporal span for 200 epochs only. Then, we define a simple baseline of a 1D blob detector~\cite{lowe2004distinctive} using the Laplacian of Gaussian kernel. To our surprise, as shown in \figref{fig:span}, this simple blob detection baseline achieves 8.2\% mAP which is already better than the Ego4D~\cite{ego4d} paper baseline of 6.0\% mAP with pretrained SlowFast~\cite{slowfast} features and VSGN~\cite{zhao2021video}, thanks to the effectiveness of Ego-Only features. We also notice that the blob detector and the frozen MAE feature prefer a longer temporal span of 16 or 32 seconds, demonstrating the importance of long-term context in egocentric videos. On the other hand, VSGN~\cite{zhao2021video} and ActionFormer~\cite{actionformer} prefer short feature spans probably because the graph neural network or the transformer captures long-term relations internally, benefiting more from local features that represent dense temporal motion. Finally, ActionFormer with finetuned features achieves the best result of 12.9\%, outperforming VSGN by 4.0\% consistently.

\begin{figure}[t]
    \centering
    \includegraphics[width=0.43\textwidth]{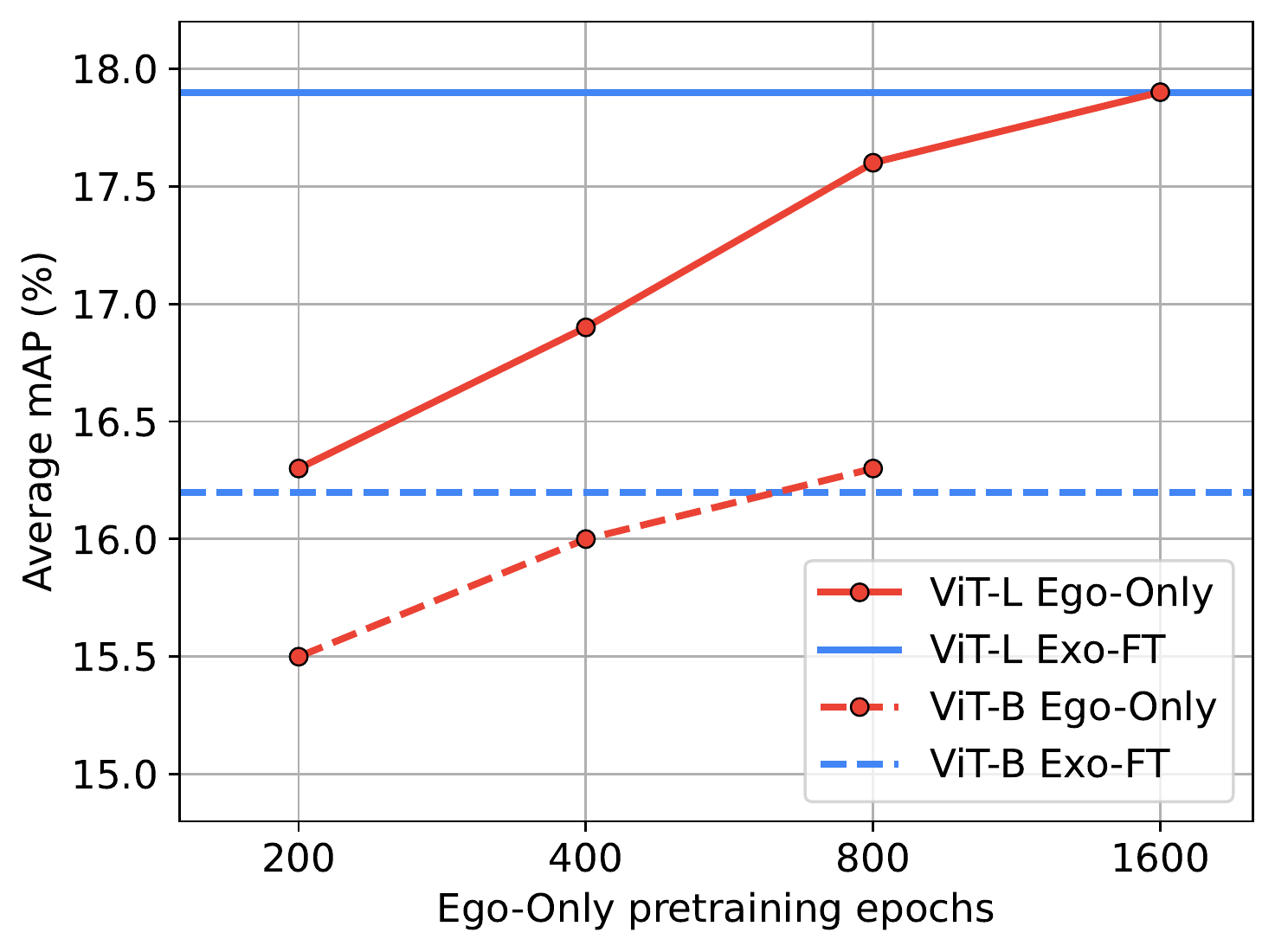}
    \caption{Scaling models and pretraining epochs. At around 800 or 1600 epochs, our Ego-Only starts to match exocentric transferring.}
    \vspace{-0.5ex}
    \label{fig:epoch}
\end{figure}

\paragraph{Scaling models and pretraining epochs.}
In addition to ablating the three stages in our Ego-Only pipeline, we also scale the model size from ViT-B to ViT-L and benchmark results under different computation budgets. We keep the relatively cheap finetuning of 20 epochs unchanged, but vary the MAE pretraining epochs. As shown in \figref{fig:epoch}, both ViT-B and ViT-L results improve consistently when they are pretrained longer. At around the budget of 800 or 1600 epochs, our Ego-Only models start to match Kinetics-400 pretrained models with both ViT-B and ViT-L. The Kinetics baselines, before transferred to egocentric data, are pretrained with 800/1600 epoch MAE and 150/100 epoch exocentric finetuning that consumes not only more data and labels but also more computation resources than Ego-Only.

\begin{table}[t]
\begin{center}
\begin{tabular}{cc|cc|c}
\toprule
ego MAE pretrain (hours) & ego finetune & mAP \\
\midrule
\color{Grey} random initialization (0h) & \color{Grey} 195h & \color{Grey} 4.2 \\
Ego4D MQ clips (195h) & 195h & 14.5 \\
Ego4D MQ videos (487h) & 195h & 14.8 \\
Ego4D EM videos (838h) & 195h & 14.7 \\
\rowcolor{Gray}
Ego4D ALL videos (3560h) & 195h & 15.5 \\
\bottomrule
\end{tabular}
\end{center}
\caption{Scaling the amount of pretraining data. {\bf MQ clips}: all MQ training clips~\cite{ego4d}. {\bf MQ videos}: all videos in the MQ task training set. {\bf EM videos}: all videos in the Episodic Memory benchmark training set. {\bf ALL videos}: all Ego4D videos except MQ val and test videos. Our Ego-Only results improve with respect to the amount of data consumed in the pretraining stage.}
\vspace{-0.5ex}
\label{tab:data}
\end{table}

\paragraph{Scaling egocentric pretraining data.}

Beyond standard ablations on pretraining epochs, an intriguing dimension for study offered by the massive scale of Ego4D is the different amounts of large-scale unsupervised video data. Specifically, given the fixed amount of finetuning data, we select four subsets and amounts of unsupervised data in Ego4D to study the data scaling property of the Ego-Only pretraining stage. Note that in all cases, we exclude val and test videos of the MQ task from the pretraining set. All models are pretrained for 200 epochs instead of 800 epochs to save computation resources. From the results in \tabref{tab:data}, we see that the performance of Ego-Only improves as more unsupervised data is provided for MAE pretraining. %

\paragraph{Scaling exocentric pretraining data.}

\begin{table}[t]
\begin{center}
\setlength\tabcolsep{3.12345pt}
\vspace{-0.5ex}
\begin{tabular}{c|ccc|ccc}
\toprule
\multirow{2}{*}{method} & self-sup. & sup. & sup. & verb & noun & \# labels \\
& MAE & exo & ego & mAP & mAP & seen \\
\midrule
\color{Grey} frozen & \color{Grey} K400 & \color{Grey} K400 & \color{Grey} - & \color{Grey} 17.9 & \color{Grey} 14.6 & \color{Grey} 307K \\
exo-FT & K400 & K400 & EPIC & 28.0 & 28.3 & 307K \\
\midrule
\color{Grey} frozen & \color{Grey} K600 & \color{Grey} K600 & \color{Grey} - & \color{Grey} 17.0 & \color{Grey} 15.0 & \color{Grey} 457K \\
exo-FT & K600 & K600 & EPIC & 27.1 & \bf 28.6 & 457K \\
\midrule
\rowcolor{Gray}
ours & EPIC & - & EPIC & \bf 29.0 & 28.1 & \bf 67K \\
\bottomrule
\end{tabular}
\end{center}
\caption{Scaling exocentric pretraining data.}
\vspace{-1.5ex}
\label{tab:k600}
\end{table}

Besides scaling egocentric data, we study the common practice of scaling exocentric pretraining from K400 (240K videos) to K600 (390K videos). As shown in \tabref{tab:k600}, scaling exocentric data improves noun mAP marginally and hurts verb mAP by 0.9\%, compared with transferring from K400. This is probably due to the bias of Kinetics towards scene and object classification.
When we evaluate on verbs, Ego-Only shows a significant absolute gain of 1.9\% over K600 transferring that requires much more labels. This observation is also consistent with the action recognition results in \tabref{tab:epic_rec}, where Ego-Only achieves the state-of-the-art verb accuracy.

\begin{table}[t]
\begin{center}
\small
\begin{tabular}{c|ccc|c}
\toprule
\multirow{2}{*}{method} & self-sup. & sup. & sup. & Ego4D \\
 & MAE & exo & ego & mAP \\
\midrule
exo-MAE & K400 & - & Ego4D & 13.4 \\
joint-MAE & K400 \& Ego4D & - & Ego4D & 16.0 \\
\rowcolor{Gray}
ours & Ego4D & - & Ego4D & \bf 16.3 \\
\bottomrule
\end{tabular}
\end{center}
\caption{Joint ego-exo pretraining.}
\vspace{-1.5ex}
\label{tab:joint_pretrain}
\end{table}

\paragraph{Joint ego-exo pretraining.} In \tabref{tab:joint_pretrain}, we study the effect of joint ego-exo pretraining by building a joint-MAE variant that trains the MAE model on both K400 and Ego4D, instead of K400 or Ego4D individually. We observe that the results are greatly improved compared with the out-of-domain K400 transferring, but lags behind our Ego-Only. 

\paragraph{Joint ego-exo finetuning.} In \tabref{tab:joint_finetune}, we explore joint ego-exo finetuning with a shared model backbone on four large-scale video datasets, including Kinetics-600, Ego4D, EPIC-Kitchens-100, and COIN~\cite{coin}. This joint dataset contains 515K labeled clips, 7$\times$ more than our default finetuning data of 67K, but does not lead to any performance gain probably due to the domain gap between these datasets.

\begin{table}[t]
\begin{center}
\setlength\tabcolsep{3.91345pt}
\begin{tabular}{c|ccc|ccc}
\toprule
\multirow{2}{*}{method} & self-sup. & sup. & sup. & verb & noun & \# labels \\
& MAE & exo & ego & mAP & mAP & seen \\
\midrule
joint-FT & EPIC & - & KEEC & 28.4 & 27.9 & 515K \\
\rowcolor{Gray}
ours & EPIC & - & EPIC & \bf 29.0 & \bf 28.1 & \bf 67K \\
\bottomrule
\end{tabular}
\end{center}
\caption{Joint ego-exo finetuning. \textbf{KEEC}: joint finetuning on \textbf{K}inetics-600, \textbf{E}go4D, \textbf{E}PIC-Kitchens-100, \textbf{C}OIN.}
\vspace{-1.5ex}
\label{tab:joint_finetune}
\end{table}

\section{Conclusion}
\label{sec:conclusion}

In this work, we have shown for the first time that we can train a state-of-the-art egocentric action detector without any exocentric transferring. Our proposed Ego-Only simplifies the current learning pipeline by removing the previous need for supervised pretraining on large-scale exocentric video or image datasets before transferring to egocentric videos. We hope our such attempt inspires the community to rethink the trade-off between training in-domain with ego-only data and transferring from out-of-domain exocentric learning. We also hope that our Ego-Only results provide a strong baseline for future research that aims to improve egocentric learning by leveraging exocentric data.

\paragraph{Acknowledgments.}
We would like to thank Christoph Feichtenhofer for sharing Video MAE code and models. We thank Effrosyni Mavroudi, Gene Byrne, Mandy Toh, Triantafyllos Afouras, Yale Song, for their advice and help.

\FloatBarrier

\appendix
\renewcommand\thefigure{\thesection.\arabic{figure}}
\renewcommand\thetable{\thesection.\arabic{table}}
\setcounter{figure}{0}
\setcounter{table}{0}
\renewcommand{\thetable}{A.\arabic{table}}

\section{Dataset Details}

\paragraph{Ego4D}~\cite{ego4d} offers 3,670 hours of daily life egocentric videos from hundreds of scenarios, providing massive-scale data for self-supervised pretraining. The Ego4D Moments Queries (MQ) task in the Episodic Memory benchmark contains 110 moments classes, 326.4 hours of videos (194.9h in train, 68.5h in val, 62.9h in test), 2522 clips (1486 in train, 521 in val, 481 in test), and 22.2K annotated temporal action segments (13.6K in train, 4.3K in val, 4.3K in test).

\paragraph{EPIC-Kitchens-100}~\cite{epic-kitchens-100} offers 100 hours (74.7h in train, 13.2h in val, 12.1h in test) of egocentric videos from 700 sessions (495 in train, 138 in val, 67 in test) in 45 kitchens. The Action Detection challenge contains 97 verb classes (97 in train, 78 in val, 84 in test), 300 noun classes (289 in train, 211 in val, 207 in test), and 90.0K temporal action segments (67.2K in train, 9.7K in val, 13.1K in test).

\paragraph{Charades-Ego}~\cite{charades-ego} offers 8K videos (3K in ego train, 3K in exo train, 846 in ego test) of daily indoor activities. The videos are recorded from both third and first person with temporal segments annotated (33K in ego train, 34K in exo train, 9K in ego test) over 157 classes.

\section{Implementation Details}
\label{sec:tech}

\paragraph{MAE pretraining.} As discussed in \secref{sec:pipeline}, we follow the technical details in video MAE~\cite{feichtenhofer2022masked} unless noted otherwise. However, as egocentric datasets contain long videos with hundreds or thousands of hours, in this paper, we define one epoch as 245,760 clips sampled from data, so that the compute budget is comparable to one Kinetics-400~\cite{kay2017kinetics} epoch. With this definition, we pretrain egocentric MAE for 800/1600 epochs, batch size 256, without repeated sampling for simplicity, learning rate 8e-4, by default. We sample clips of 16 frames with a temporal span of 2 seconds, equivalent to a sampling rate of 4 in 30-fps videos.

\paragraph{Finetuning.} We finetune for 20 epochs with 2-epoch warm-up, batch size 128, RandAugment~\cite{cubuk2020randaugment}, stochastic depth~\cite{huang2016deep} 0.2, dropout~\cite{srivastava2014dropout} 0.5, label smoothing 0.0001 for BCE, no mixup~\cite{zhang2017mixup} or cutmix~\cite{yun2019cutmix} as they are not common for segmentation. We use SGD with learning rate 4.0 weight decay 0.0 on Ego4D, while we use AdamW~\cite{adamw} with learning rate 8e-4 weight decay 0.05 on EPIC-Kitchens-100. For finetuning on EPIC-Kitchens-100, we concatenate all verb and noun classes so that we finetune only once.

\paragraph{Action detection.} As discussed in \secref{sec:pipeline}, we follow the details of ActionFormer~\cite{actionformer} for EPIC-Kitchens-100 unless noted otherwise. Our Ego4D features are extracted at stride 8 which equals the transformer output stride, with frame sampling rate 4 and temporal patch stride 2. The sliding windows use stride 8 as well. We train for 10 epochs with 8-epoch warm-up, learning rate 2e-4. EPIC-Kitchens-100 features use stride 16~\cite{actionformer} for fair comparison. We train for 20 epochs with 16-epoch warm-up, learning rate 2e-4. We report an average of 3 runs.

\paragraph{Action recognition.} We sample clips of 32 frames~\cite{memvit} with a temporal span of 3.2 seconds, equivalent to a sampling rate of 3 in 30-fps videos. And due to the extra memory constraint, we reduce the batch size to 64. On Charades-Ego without exocentric data, we train 10 epochs with 1-epoch warm up, SGD optimizer, learning rate 0.8, and no weight decay. On Charades-Ego with exocentric data, we instead use AdamW optimizer, learning rate 2.4e-4, and weight decay 0.05. On EPIC-Kitchens-100, we train 20 epochs with 2-epoch warm up, AdamW optimizer, learning rate 2.4e-4, and weight decay 0.05.

\begin{figure}[!ht]
    \centering
    \includegraphics[width=0.45\textwidth]{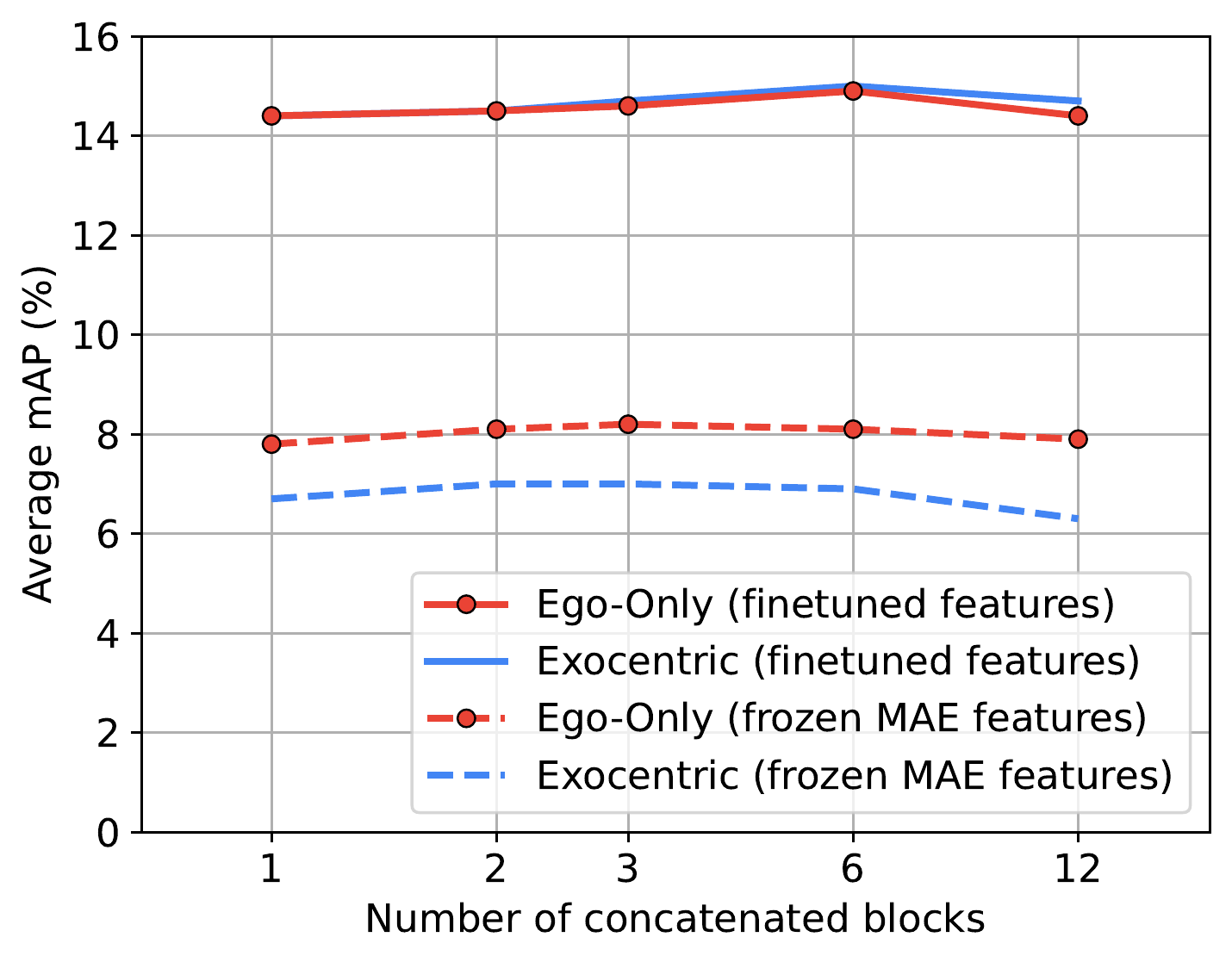}
    \caption{Ego4D Moments Queries results with concatenated features from the last few (2, 3, 6, 12) transformer blocks (12 blocks in total for the ViT-B~\cite{dosovitskiy2020image} architecture), instead of our default choice of the last block only. The detection results are almost not affected in any of the four models studied. This stable gap between finetuned features and frozen MAE features verifies the necessity of the egocentric finetuning stage in Ego-Only.}
    \label{fig:concat}
\end{figure}

\section{Ablation on Concatenated Features}
\label{sec:concat}
In \figref{fig:concat}, we present the ablation of concatenating features from the last few (2, 3, 6, or 12) transformer blocks, instead of our default choice of the last block only. This is inspired by the linear protocol in DINO\cite{dino} that was aimed to improve results with frozen self-supervised learning features (in our case frozen MAE features) but we ablate this choice for all models, with and without finetuning. However, we see a marginal gain for frozen MAE features, which confirms the necessity of the egocentric finetuning stage in Ego-Only.

\begin{table}[t]
\begin{center}
\setlength\tabcolsep{2.02345pt}
\begin{tabular}{l|cc|cc}
\toprule
method & MAE & exo & rebalancing technique & mAP \\
\midrule
exo-FT & K400 & K400 & resampling & 16.2 \\
exo-FT & K400 & K400 & per-class reweighting & 14.4 \\
\rowcolor{Gray}
exo-FT & K400 & K400 & per-instance reweighting & 16.2 \\
\midrule
\small Ego-Only & \small Ego4D & - & resampling & 16.3 \\
\small Ego-Only & \small Ego4D & - & per-class reweighting & 14.4 \\
\rowcolor{Gray}
\small Ego-Only & \small Ego4D & - & per-instance reweighting & 16.3 \\
\bottomrule
\end{tabular}
\end{center}
\caption{Varying rebalancing techniques. Ego-Only matches exocentric transferring regardless of rebalancing techniques.}
\label{tab:rebalancing}
\end{table}

\section{Ablation on Rebalancing Techniques.}

As discussed in \secref{sec:finetune}, we are currently mitigating the imbalance challenges by simply reweighting the loss according to the number of positive frames in each action instance. Beyond this current technique, we also study a simple action resampling option as a natural alternative. Specifically, instead of uniformly sampling all the clips within the train data, we sample only the center 2 seconds of each action regardless of the action length, similar to an action classification task. As shown in \tabref{tab:rebalancing}, this resampling option performs the same as the default reweighting with and without exocentric transferring. We also study a per-class reweighting method that ignores action length imbalance within a class and find that it performs worse than the other two rebalancing methods. In all these cases, our Ego-Only method matches Kinetics transferring, without any exocentric data or label, and {\it regardless} of the rebalancing techniques employed. We consider further exploration of better rebalancing methods as an open research problem and leave it to future work beyond the scope of this paper.

\section{Error Analyses}
\label{sec:analysis}

\begin{figure*}[t]
    \centering
    \begin{tabular}{c|c|c}
        \toprule
        EPIC & Exocentric transferring (previous) & Ego-Only (ours) \\
        \midrule
         & 25.6\% mAP & 27.7\% mAP \\[.3ex]
        Verb &
        \raisebox{-.5\totalheight}{
        \includegraphics[width=0.4\textwidth]{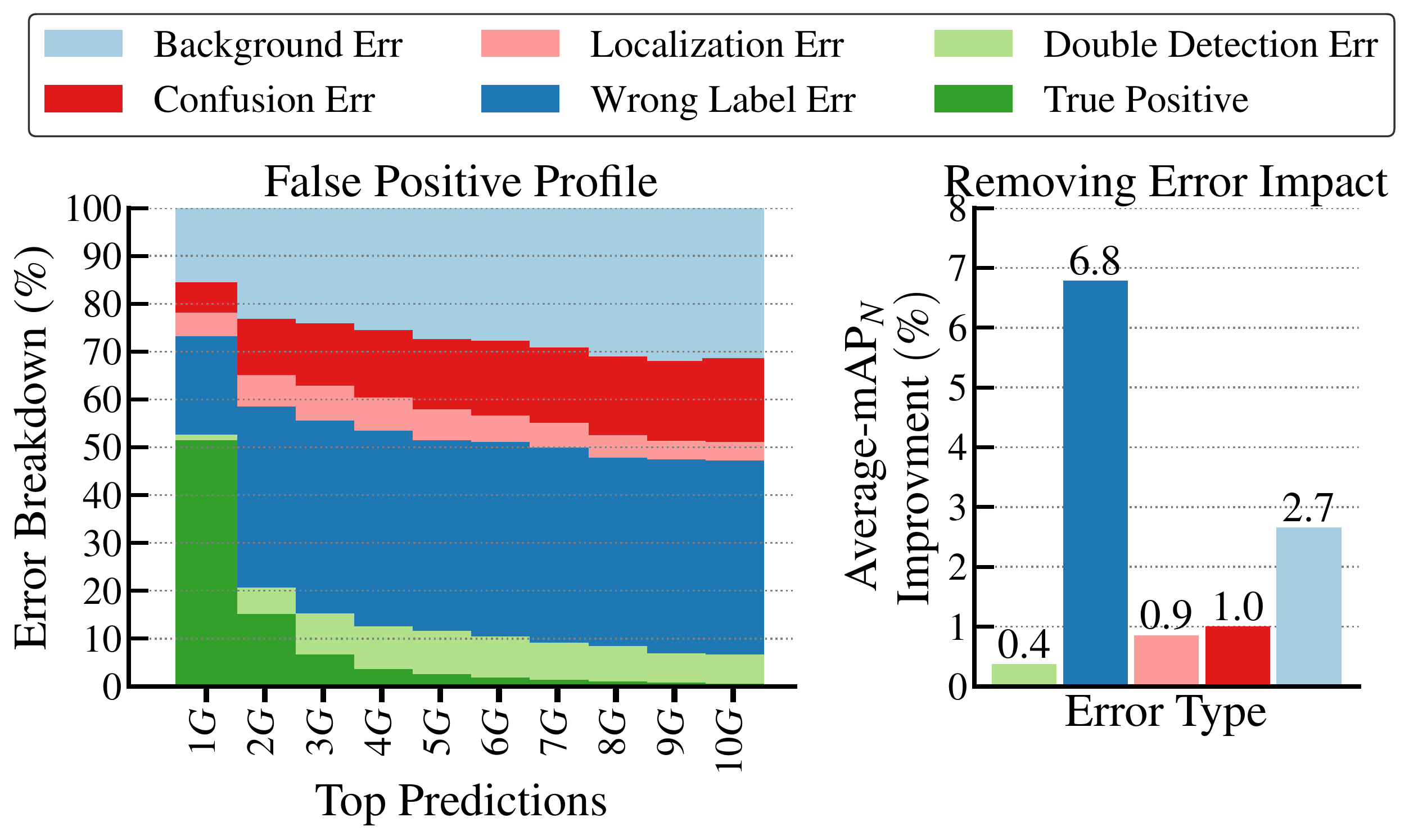}} &
        \raisebox{-.5\totalheight}{
        \includegraphics[width=0.4\textwidth]{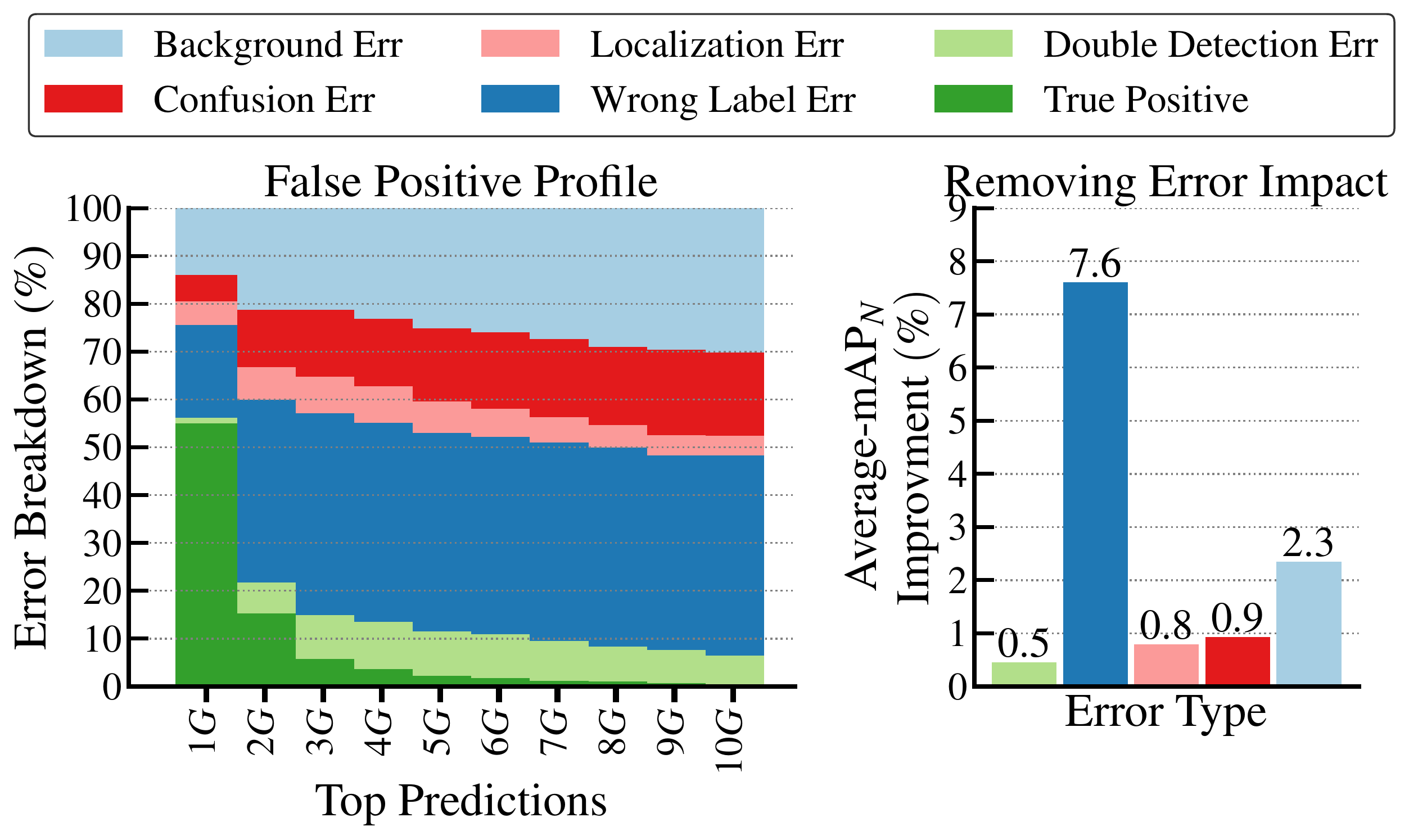}} \\
        \midrule
         & 26.3\% mAP & 28.1\% mAP \\[.3ex]
        Noun &
        \raisebox{-.5\totalheight}{
        \includegraphics[width=0.4\textwidth]{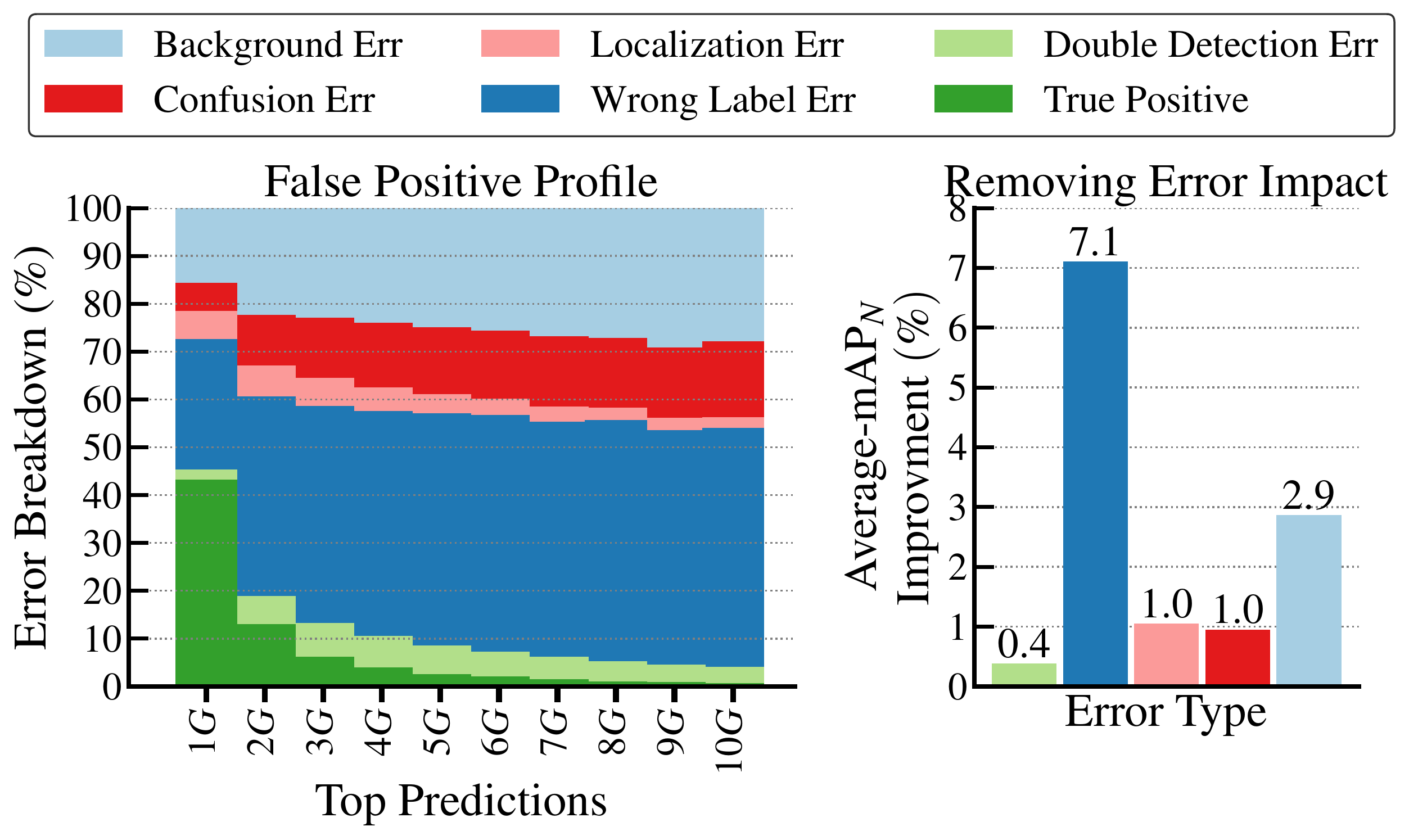}} &
        \raisebox{-.5\totalheight}{
        \includegraphics[width=0.4\textwidth]{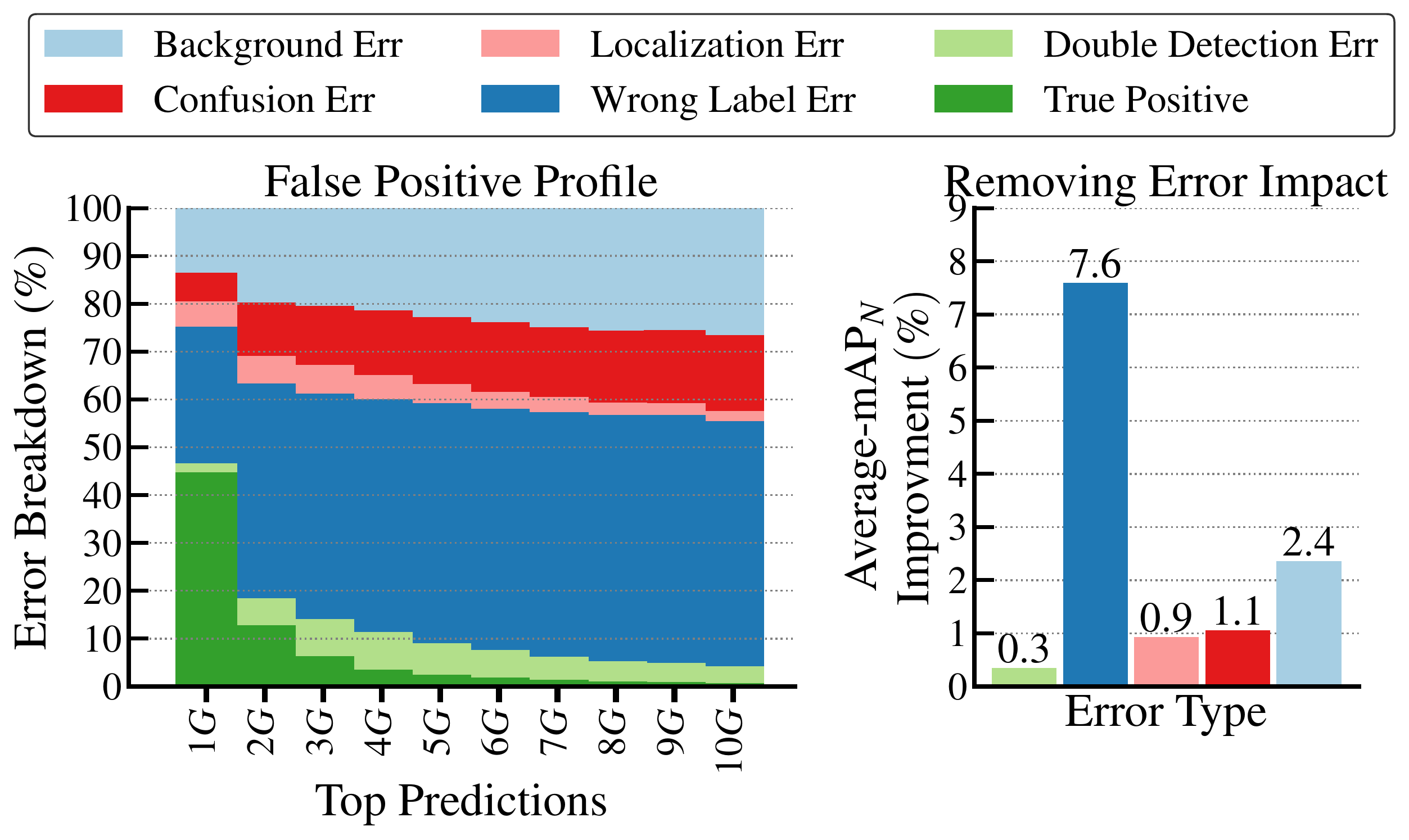}} \\
        \bottomrule
    \end{tabular}
    \vspace{2ex}
    \caption{False positive analysis on EPIC-Kitchens-100~\cite{epic-kitchens-100} with DETAD~\cite{alwassel_2018_detad}. The error types are determined by the tIoU between ground-truth and predicted segments, as well as the correctness of the predicted labels. Background error: tIoU $<$ 1e-5; confusion error: 1e-5 $<$ tIoU $< \alpha$ and label is wrong; localization error: label is correct but 1e-5 $<$ tIoU $<\alpha$; wrong label error: tIoU $>=\alpha$ but label is wrong, where $\alpha$ refers to the tIoU thresholds \{0.1, 0.2, 0.3, 0.4, 0.5\}. `G' refers to the number of ground-truth instances. According to the error breakdown, although the large-scale exocentric pretraining helps reducing wrong label errors, our Ego-Only predicts more true positives correctly and reduces background errors, probably because Kinetics~\cite{kay2017kinetics} contains mostly trimmed videos with foreground actions only.}
    \label{fig:false_positive}
\end{figure*}

\paragraph{False positive analysis.}
In \figref{fig:false_positive}, we analyze false positive errors on EPIC-Kitchens-100~\cite{epic-kitchens-100} with ViT-L~\cite{dosovitskiy2020image} models using the DETAD~\cite{alwassel_2018_detad} error diagnosing tool. The models are trained with per-class reweighting. We notice that Ego-Only reduces false positive errors on backgrounds, compared with exocentric pretraining baselines, probably because Kinetics~\cite{kay2017kinetics} contains mostly trimmed videos with foreground actions only.

\begin{figure*}[t]
    \centering
    \begin{tabular}{c|c|c}
        \toprule
        EPIC & Exocentric transferring (previous) & Ego-Only (ours) \\
        \midrule
         & 25.6\% mAP & 27.7\% mAP \\[.3ex]
        Verb &
        \raisebox{-.5\totalheight}{
        \includegraphics[width=0.4\textwidth]{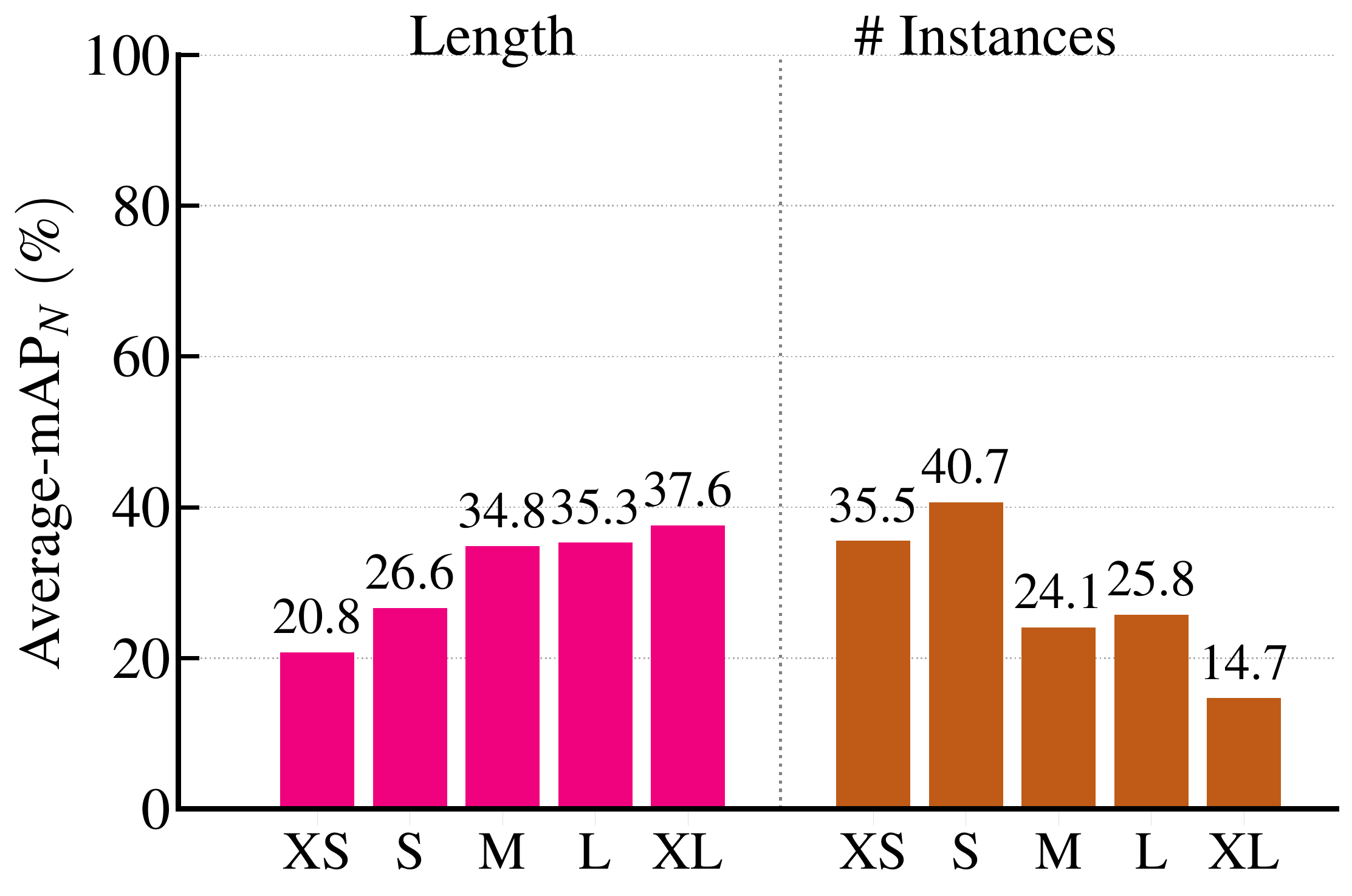}} &
        \raisebox{-.5\totalheight}{
        \includegraphics[width=0.4\textwidth]{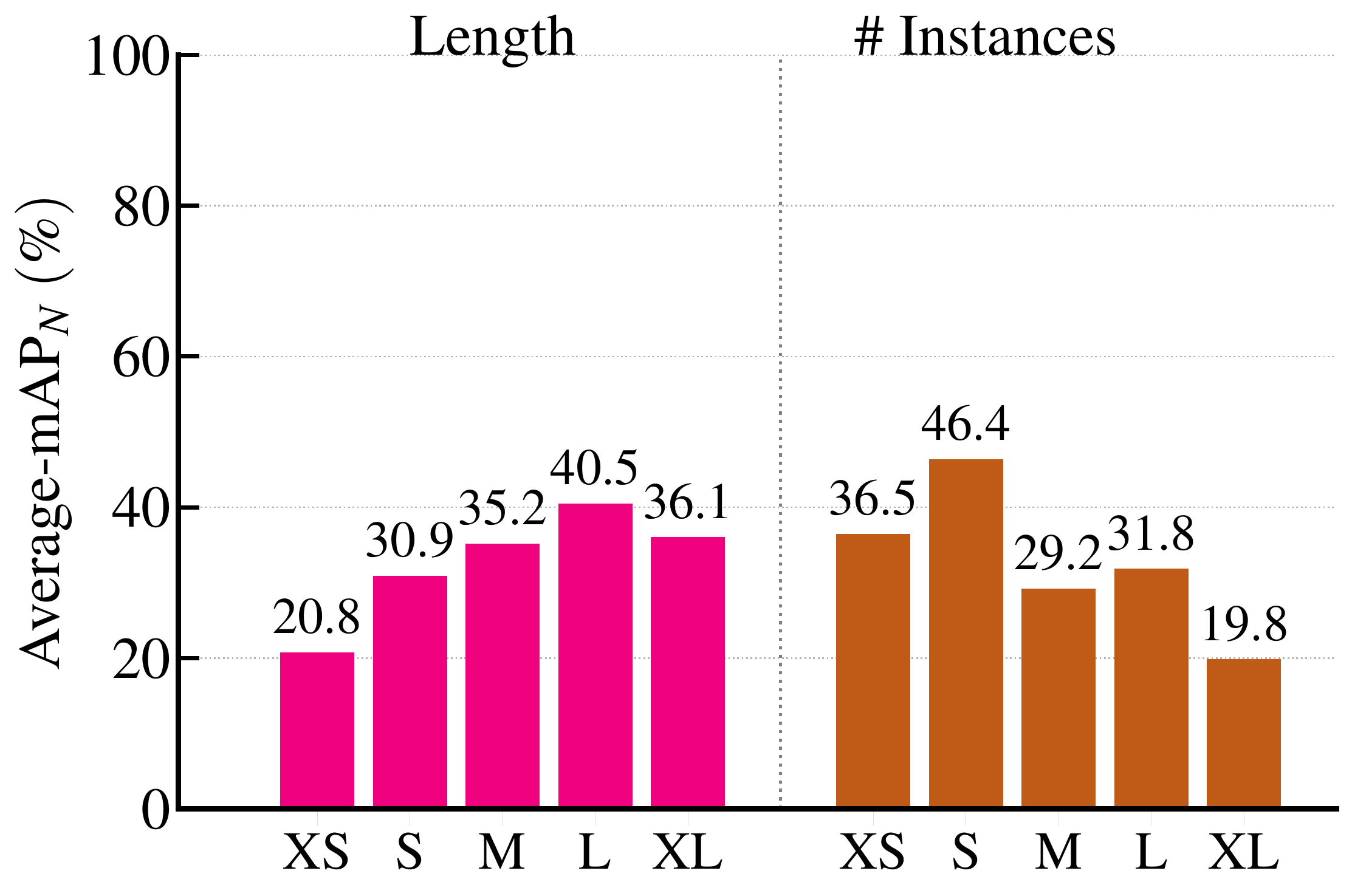}} \\
        \midrule
         & 26.3\% mAP & 28.1\% mAP \\[.3ex]
        Noun &
        \raisebox{-.5\totalheight}{
        \includegraphics[width=0.4\textwidth]{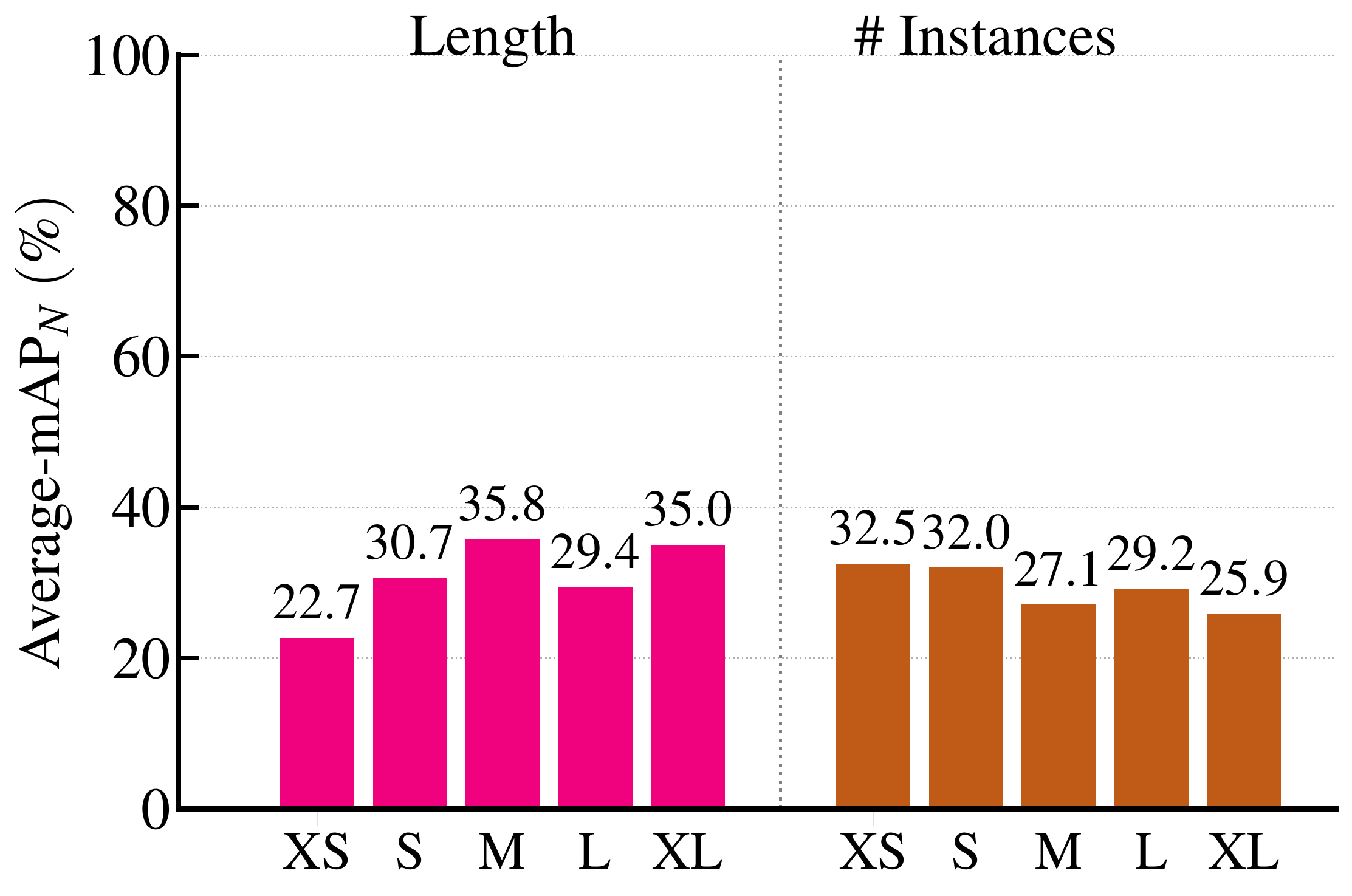}} &
        \raisebox{-.5\totalheight}{
        \includegraphics[width=0.4\textwidth]{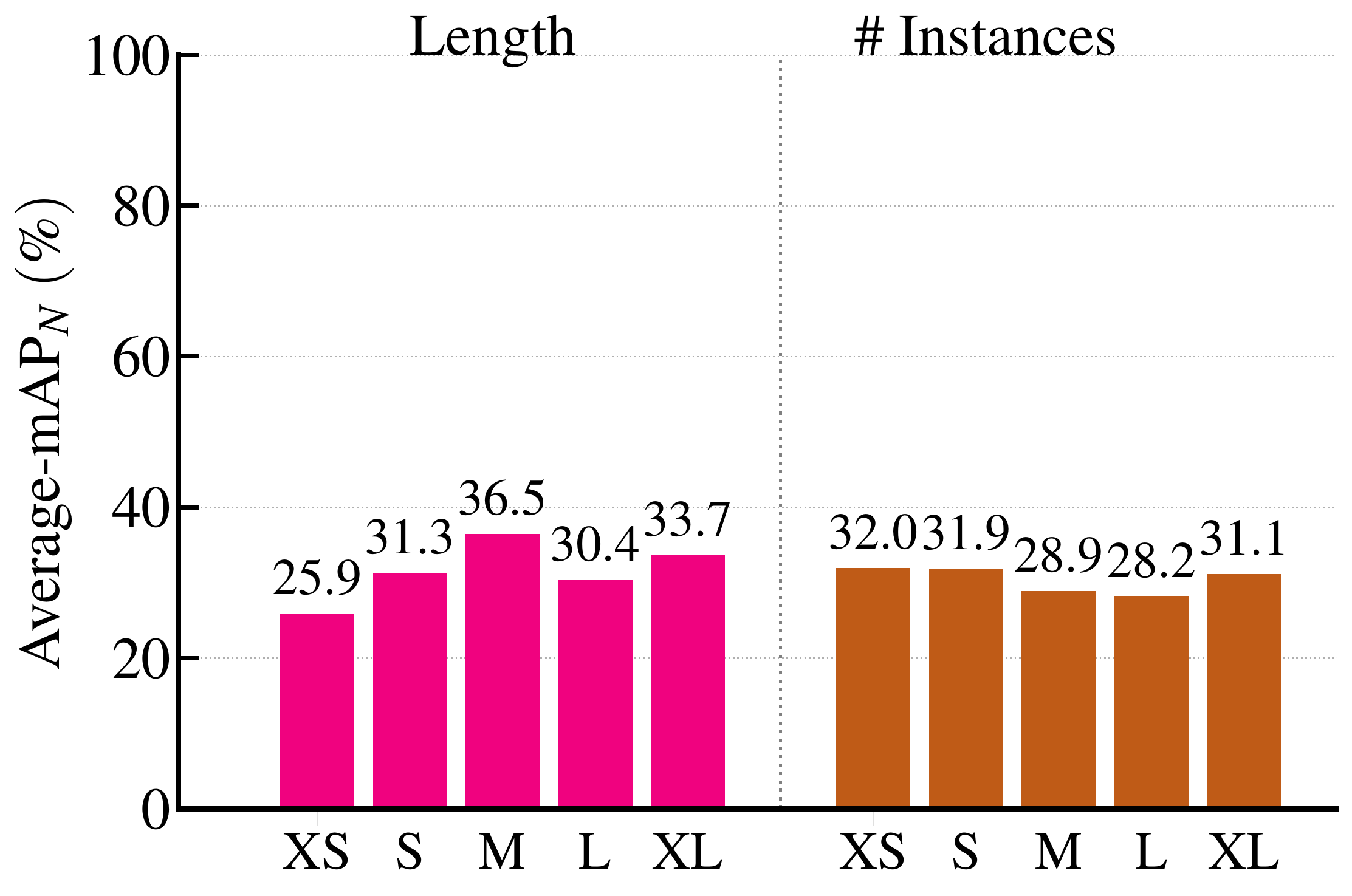}} \\
        \bottomrule
    \end{tabular}
    \vspace{2ex}
    \caption{Sensitivity analysis on EPIC-Kitchens-100~\cite{epic-kitchens-100} with DETAD~\cite{alwassel_2018_detad}. Ground-truth segments are divided into 5 equal buckets according to their characteristic~\cite{alwassel_2018_detad} percentiles. Then, average mAP\textsubscript{N}~\cite{alwassel_2018_detad} metrics are computed for each characteristic bucket. The `length' characteristic measures the length of the ground-truth action segment in seconds. The `\# instances' characteristic measures the number of action instances belonging to the same category as the ground-truth segment in the same video. According to the average mAP\textsubscript{N} in each bucket, we observe that our Ego-Only improves significantly when there are multiple verb instances of the same category in a video.}
    \label{fig:sensitivity}
\end{figure*}

\paragraph{Sensitivity analysis.}
In \figref{fig:sensitivity}, we analyze the model sensitivity according to DETAD characteristics~\cite{alwassel_2018_detad} on EPIC-Kitchens-100~\cite{epic-kitchens-100} with ViT-L~\cite{dosovitskiy2020image} models. The models are trained with per-class reweighting. We observe that our Ego-Only improves significantly when there are multiple verb instances of the same category in a video.

\section{Visualization of MAE Reconstructions}
\begin{figure*}[t]
    \centering
    \begin{tabular}{cc}
        \includegraphics[width=0.47\textwidth]{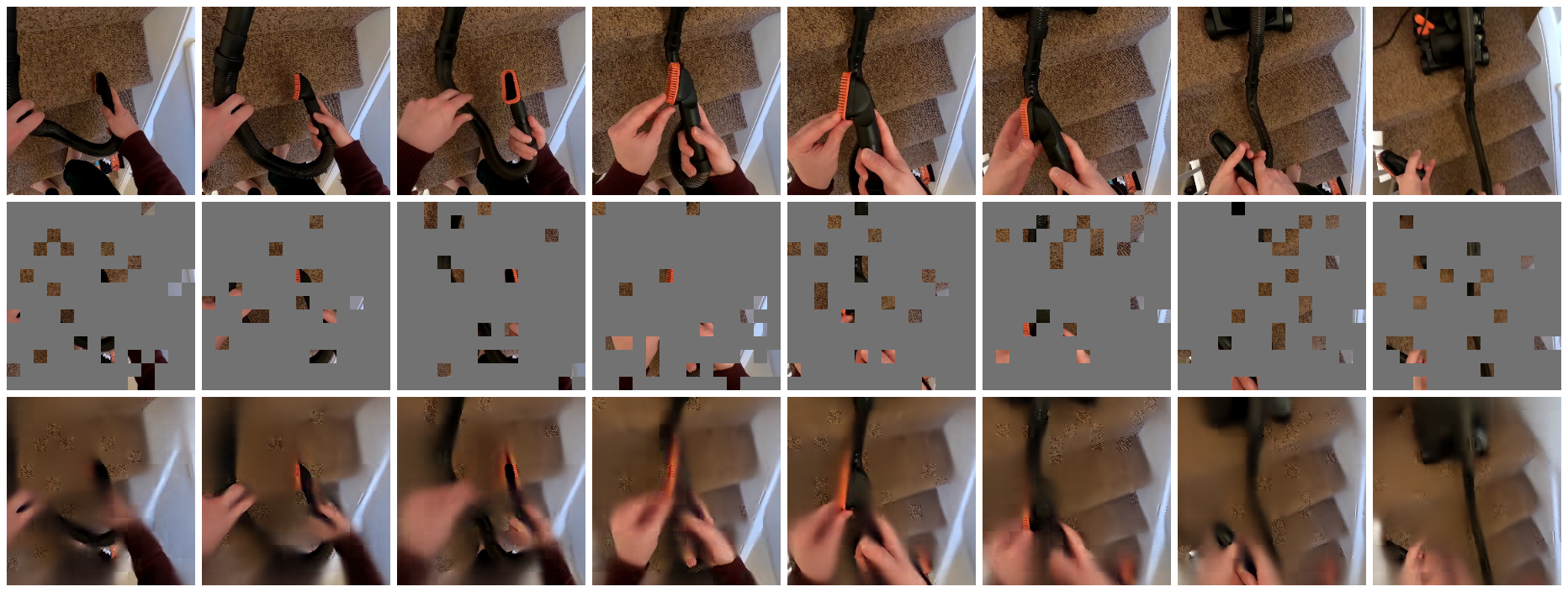} &
        \includegraphics[width=0.47\textwidth]{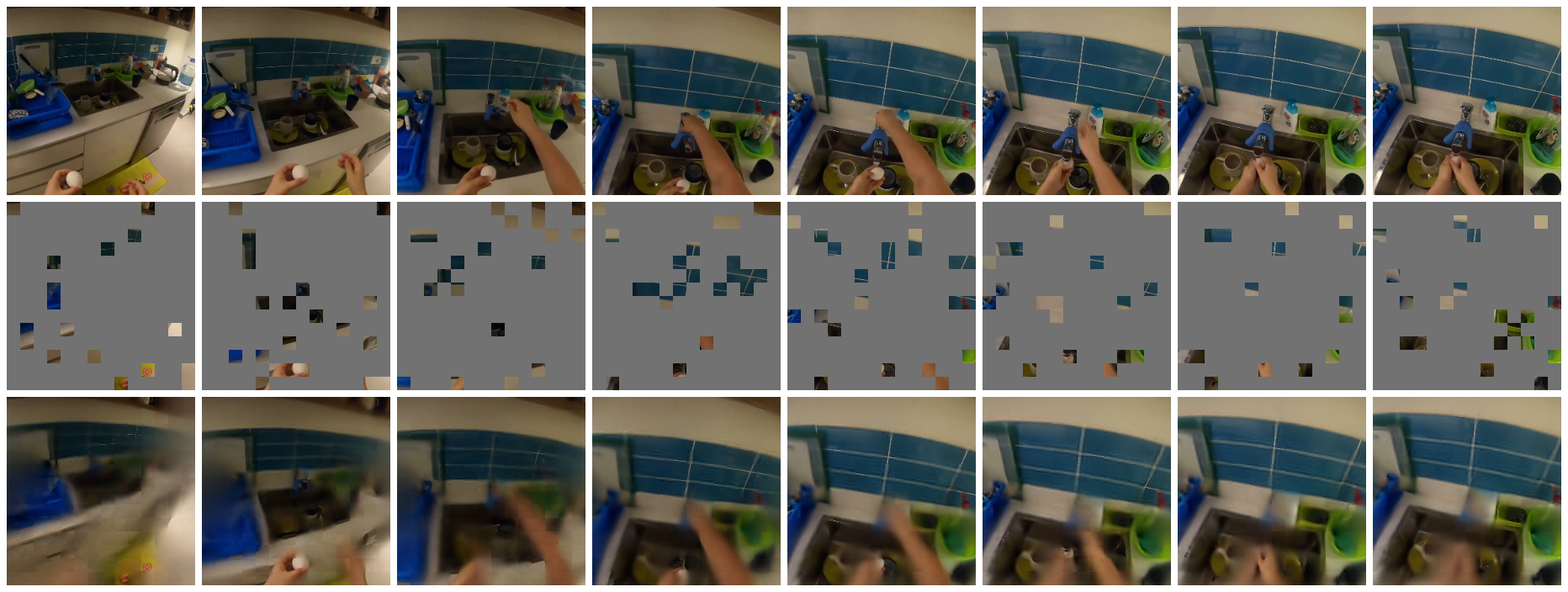} \\
        (a) & (b) \\
        \includegraphics[width=0.47\textwidth]{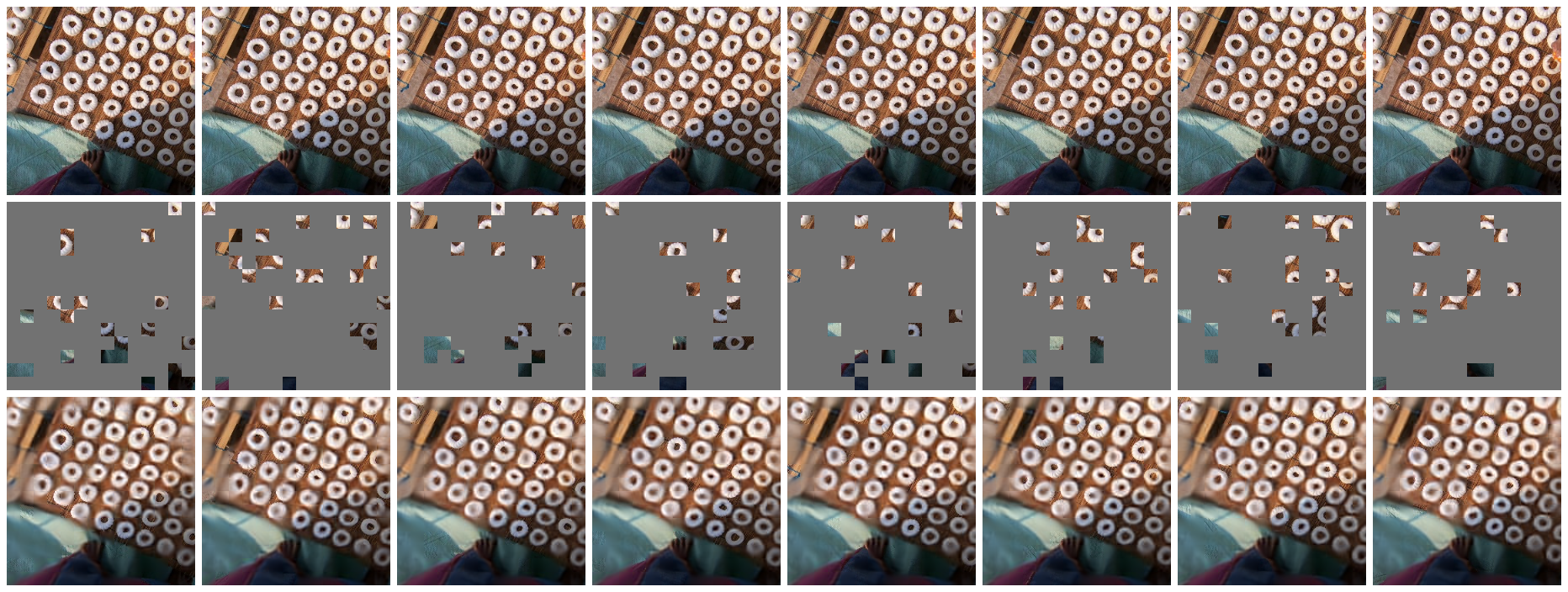} &
        \includegraphics[width=0.47\textwidth]{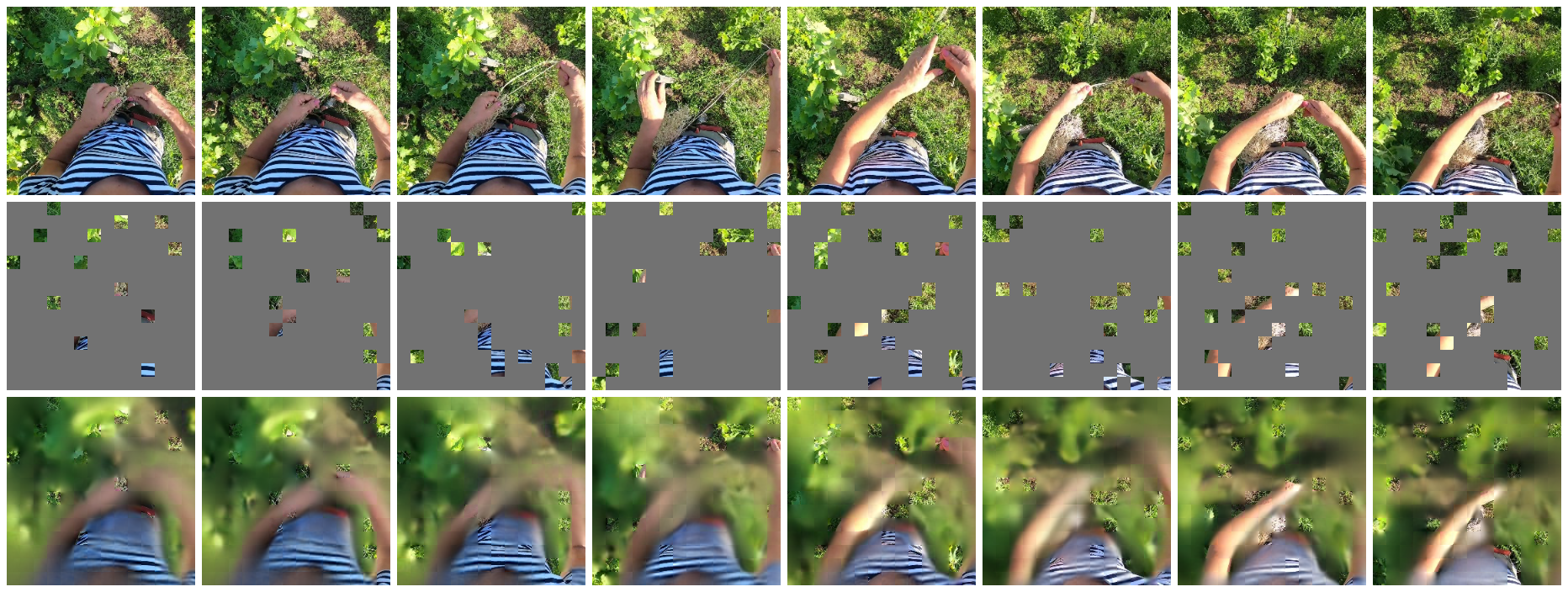} \\
        (c) & (d) \\
        \includegraphics[width=0.47\textwidth]{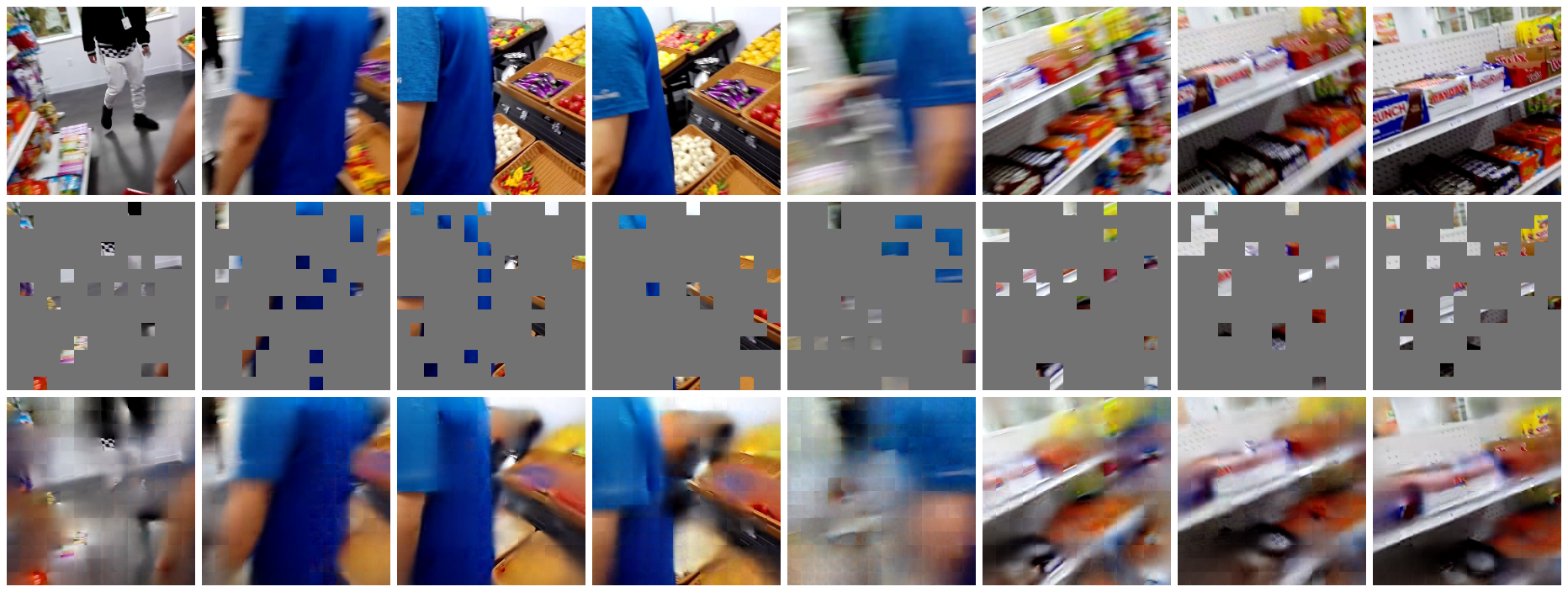} &
        \includegraphics[width=0.47\textwidth]{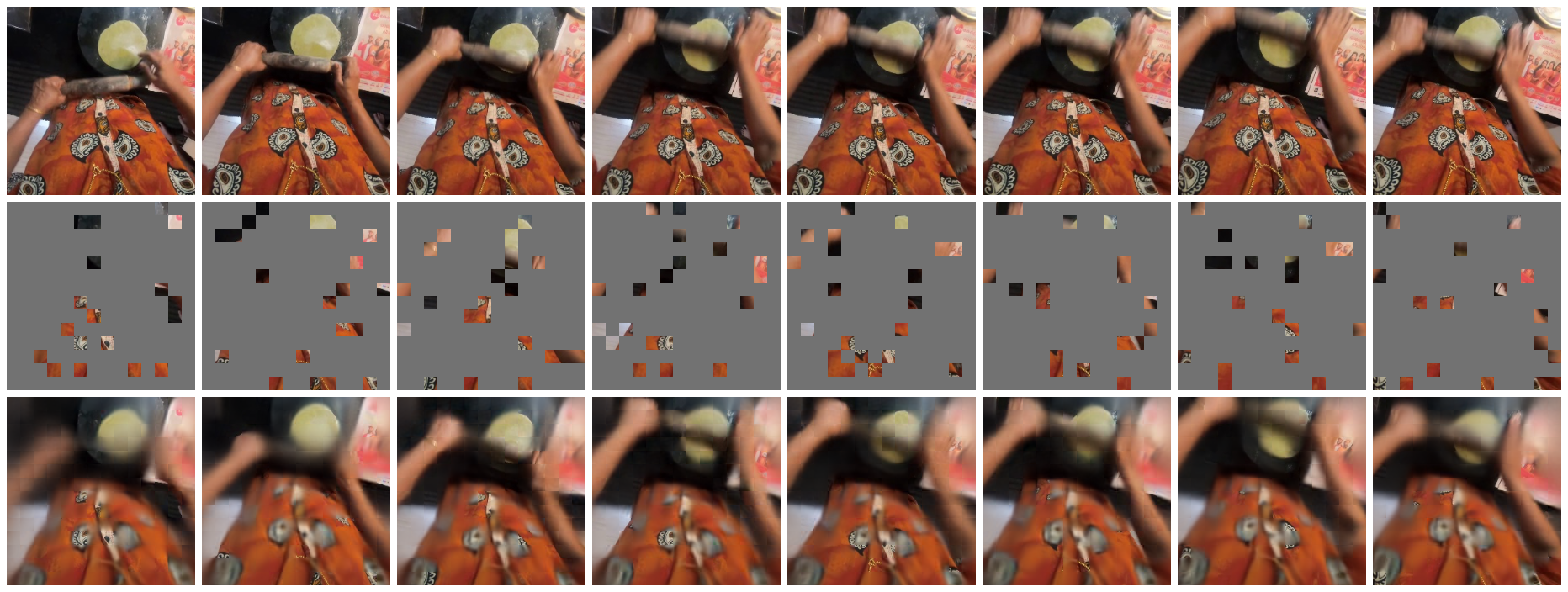} \\
        (e) & (f) \\
        \includegraphics[width=0.47\textwidth]{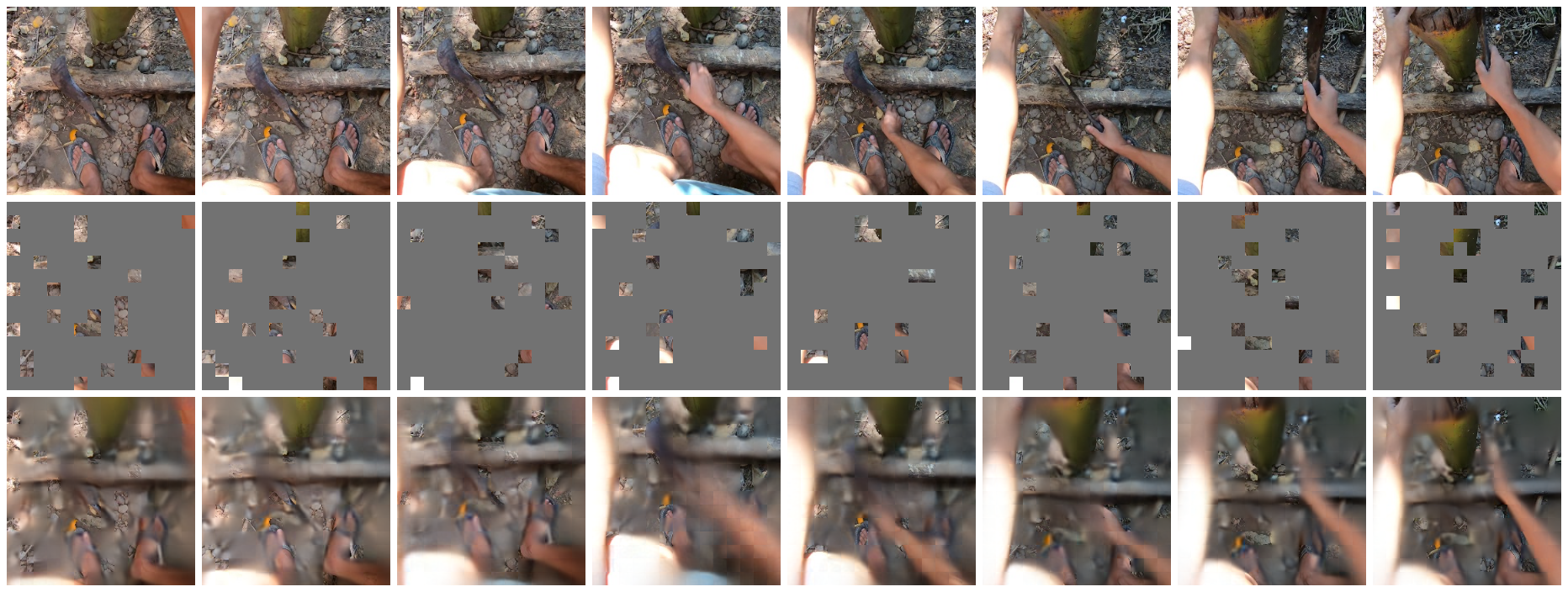} &
        \includegraphics[width=0.47\textwidth]{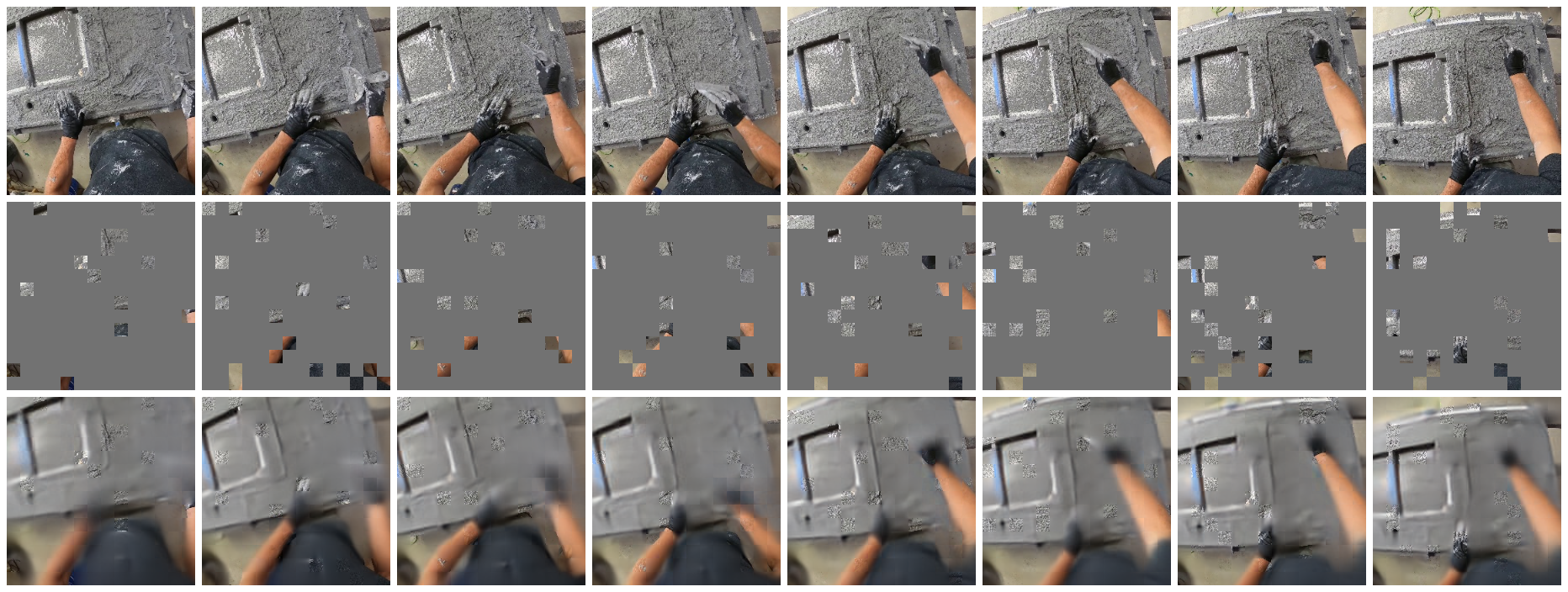} \\
        (g) & (h) \\
        \includegraphics[width=0.47\textwidth]{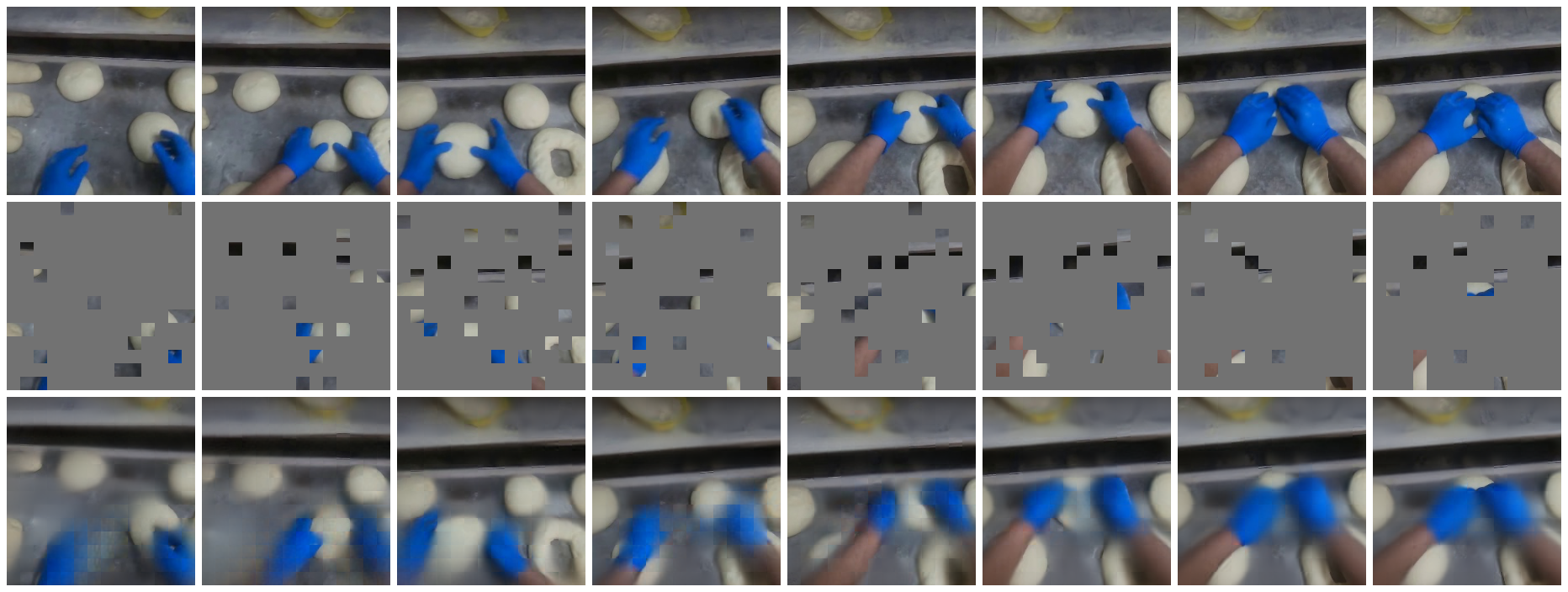} &
        \includegraphics[width=0.47\textwidth]{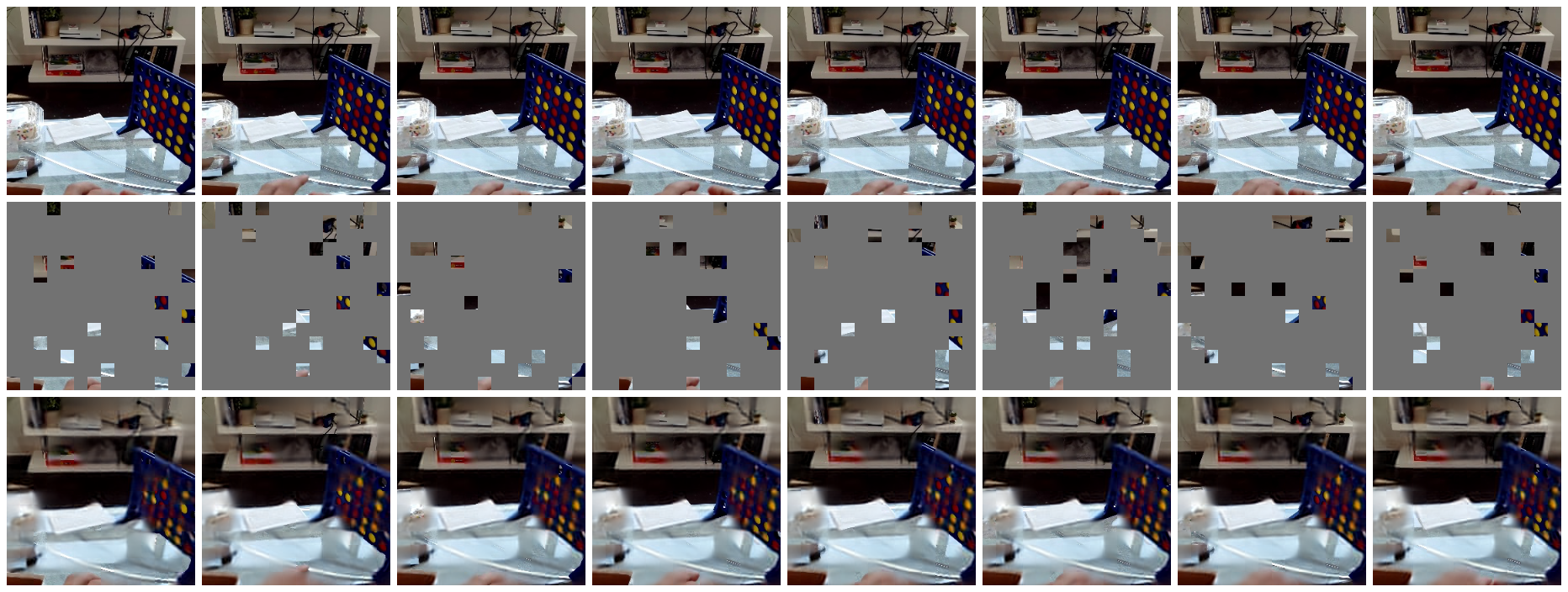} \\
        (i) & (j) \\
        \includegraphics[width=0.47\textwidth]{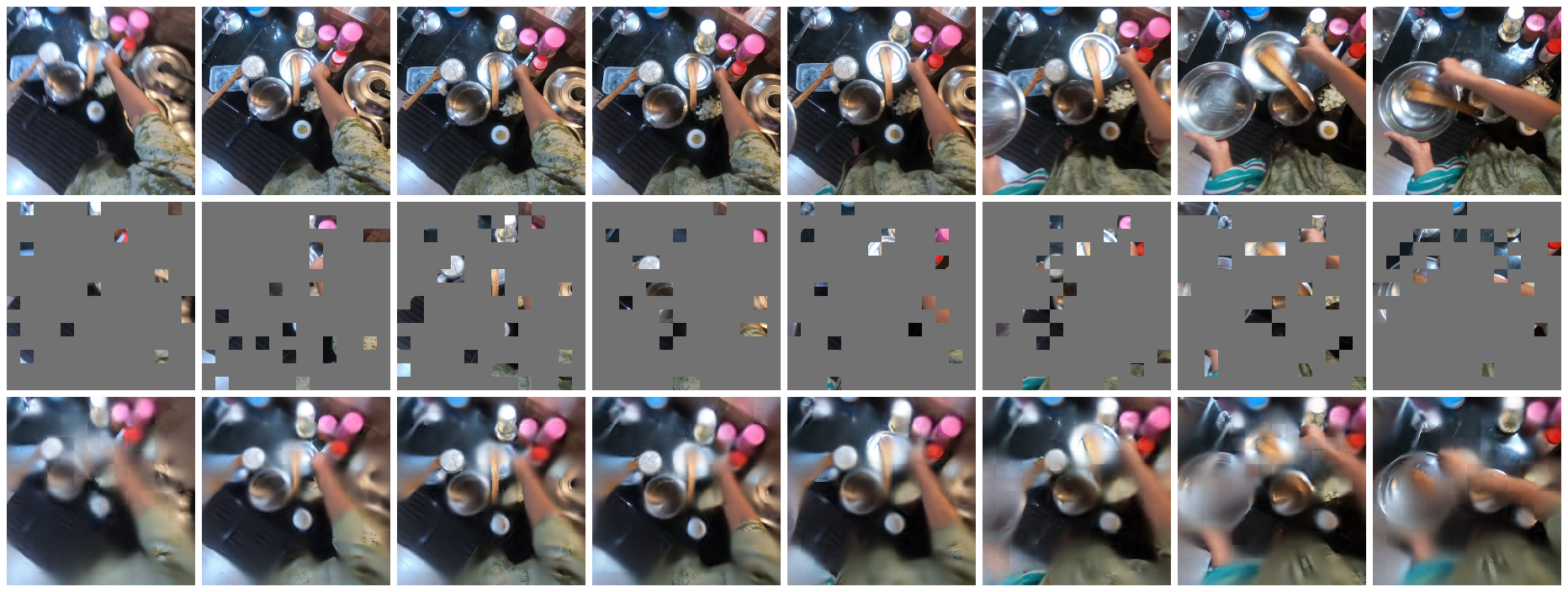} &
        \includegraphics[width=0.47\textwidth]{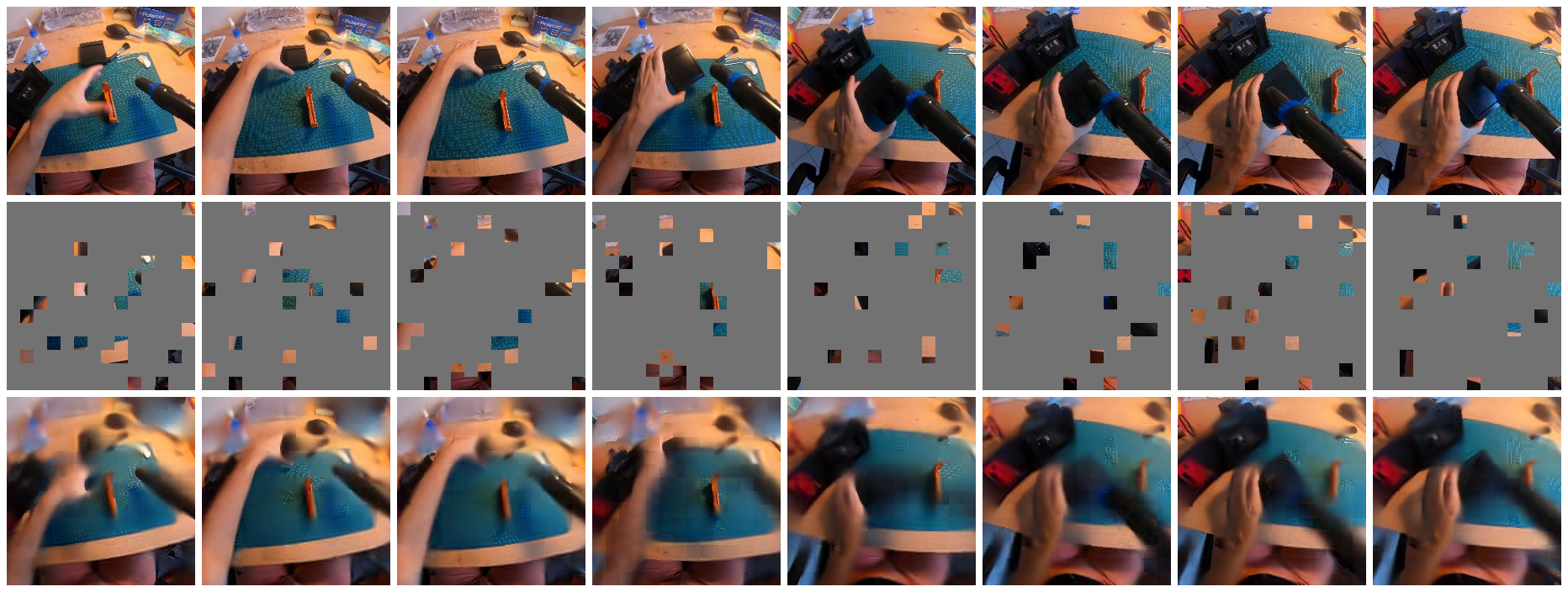} \\
        (k) & (l) \\
    \end{tabular}
    \caption{MAE~\cite{mae,feichtenhofer2022masked} reconstruction results on Ego4D~\cite{ego4d} MQ {\it val} set. For each sample, we show the original video (top), the randomly masked video (middle), and the MAE reconstruction (bottom). We visualize 8 frames~\cite{feichtenhofer2022masked} out of 16 with a temporal stride of 2. The model predicts RGB pixels without patch normalization with a masking ratio of 90\%. We notice that egocentric MAE learns human-object interactions (d,f,g,h,i,k) and temporal correspondence across frames (c,j), even in cases with strong head/camera motion (a,b,e,l).}
    \label{fig:vis}
\end{figure*}

In \figref{fig:vis}, we visualize the MAE~\cite{mae,feichtenhofer2022masked} reconstruction results on a few Ego4D~\cite{ego4d} examples with a ViT-B~\cite{dosovitskiy2020image} trained for 200 epochs without per-patch normalization. We notice that egocentric MAE learns human-object interactions (d,f,g,h,i,k) and temporal correspondence across frames (c,j), even in cases with strong head/camera motion (a,b,e,l).

\renewcommand{\arraystretch}{0.5}

\foreach \x in {1,...,123}{\phantom{x}\\}
\FloatBarrier
{\small
\bibliographystyle{ieee_fullname}
\bibliography{egbib}
}

\end{document}